\documentclass[Afour,sageh,times]{sagej}

\usepackage{amsmath}
\usepackage{amssymb}
\usepackage{bm}
\usepackage{graphicx}
\usepackage{booktabs}
\usepackage{multirow}
\usepackage[table]{xcolor}
\usepackage{makecell}

\usepackage[inline]{enumitem}
\usepackage{lipsum}
\usepackage{titlesec}
\usepackage{wasysym}

\usepackage[
pagebackref,
breaklinks,
bookmarksopen,
bookmarksnumbered,
colorlinks,
citecolor=blue,
urlcolor=blue,
linkcolor=blue
]{hyperref}
\usepackage[capitalize]{cleveref}
\crefname{section}{Sec.}{Secs.}
\Crefname{section}{Section}{Sections}
\crefname{table}{Tab.}{Tabs.}
\Crefname{table}{Table}{Tables}

\usepackage{tcolorbox}
\tcbuselibrary{breakable, skins}
\newtcolorbox{promptbox}[1][]{
  breakable,
  colback=gray!5,
  colframe=gray!40,
  boxrule=0.4pt,
  arc=2pt,
  left=6pt, right=6pt, top=6pt, bottom=6pt,
  fontupper=\small\ttfamily,
  #1
}

\usepackage{listings}
\newcommand{\ie}{\textit{i.e.}}
\newcommand{\eg}{\textit{e.g.}}

\newcommand{\systemname}{DYMO-Hair}

\newlength\savewidth
\newcommand\shline{\noalign{\global\savewidth\arrayrulewidth\global\arrayrulewidth 1pt}\hline\noalign{\global\arrayrulewidth\savewidth}}

\begin{document}

\runninghead{Zhao et al.}

\title{Learning Latent Volumetric Dynamics for Multimodal Goal-Conditioned Robot Manipulation of Hair}

\author{Chengyang Zhao\affilnum{1}, 
Uksang Yoo\affilnum{1}, 
Arkadeep Narayan Chaudhury\affilnum{2}, 
Giljoo Nam\affilnum{3}, 
Nathan Dennler\affilnum{4},
Andreea Bobu\affilnum{4},
Jonathan Francis\affilnum{1,5}, 
Jeffrey Ichnowski\affilnum{1}, 
Jean Oh\affilnum{1}}

\affiliation{\affilnum{1}Robotics Institute, Carnegie Mellon University \\
\affilnum{2}Apple \\
\affilnum{3}Meta \\
\affilnum{4}Massachusetts Institute of Technology \\
\affilnum{5}Bosch Center for Artificial Intelligence}

\corrauth{Jean Oh, 
Robotics Institute, 
Carnegie Mellon University, 
5000 Forbes Ave, 
Pittsburgh, 
PA 15213, 
USA}

\email{jeanoh@cmu.edu}

\begin{abstract}

Hair care is an essential daily activity for many, yet it remains inaccessible to individuals with limited mobility and challenging for robot systems due to the fine-grained physical structure and complex dynamics of hair. 
In this work, we present \systemname, a model-based robot hair care system for generalizable multimodal goal-conditioned hair styling. 
We introduce a volumetric dynamics learning method that models combing-induced hair deformations through action-conditioned editing in a pre-trained hierarchical 3D latent space. This latent space captures diverse hair states through volumetric occupancy and local strand orientations, while large-scale pre-training, supported by a physics-based hair-combing simulator, enables generalization to previously unseen hairstyles.
Coupled with a Model Predictive Path Integral (MPPI) planner, the learned dynamics model enables closed-loop hair styling toward both visual references and natural language instructions as user input.
 In closed-loop experiments, \systemname{} achieves lower geometric error than baseline systems for visual goals and receives significantly higher ratings than VLM-based planners for language goals in a 58-participant user study. 
Real-world experiments further demonstrate zero-shot transfer to seven physical wigs varying in color, length, and texture, effectively styling hair toward both visual and language goals without real-world fine-tuning.
Together, these results represent a step toward more generalizable, flexible, and accessible robot hair styling in the real world.
More details are available on: \href{https://dymohair.github.io/}{dymohair.github.io}.

\end{abstract}


\maketitle

\begin{figure*}[t]
    \centering
    \includegraphics[width=\linewidth]{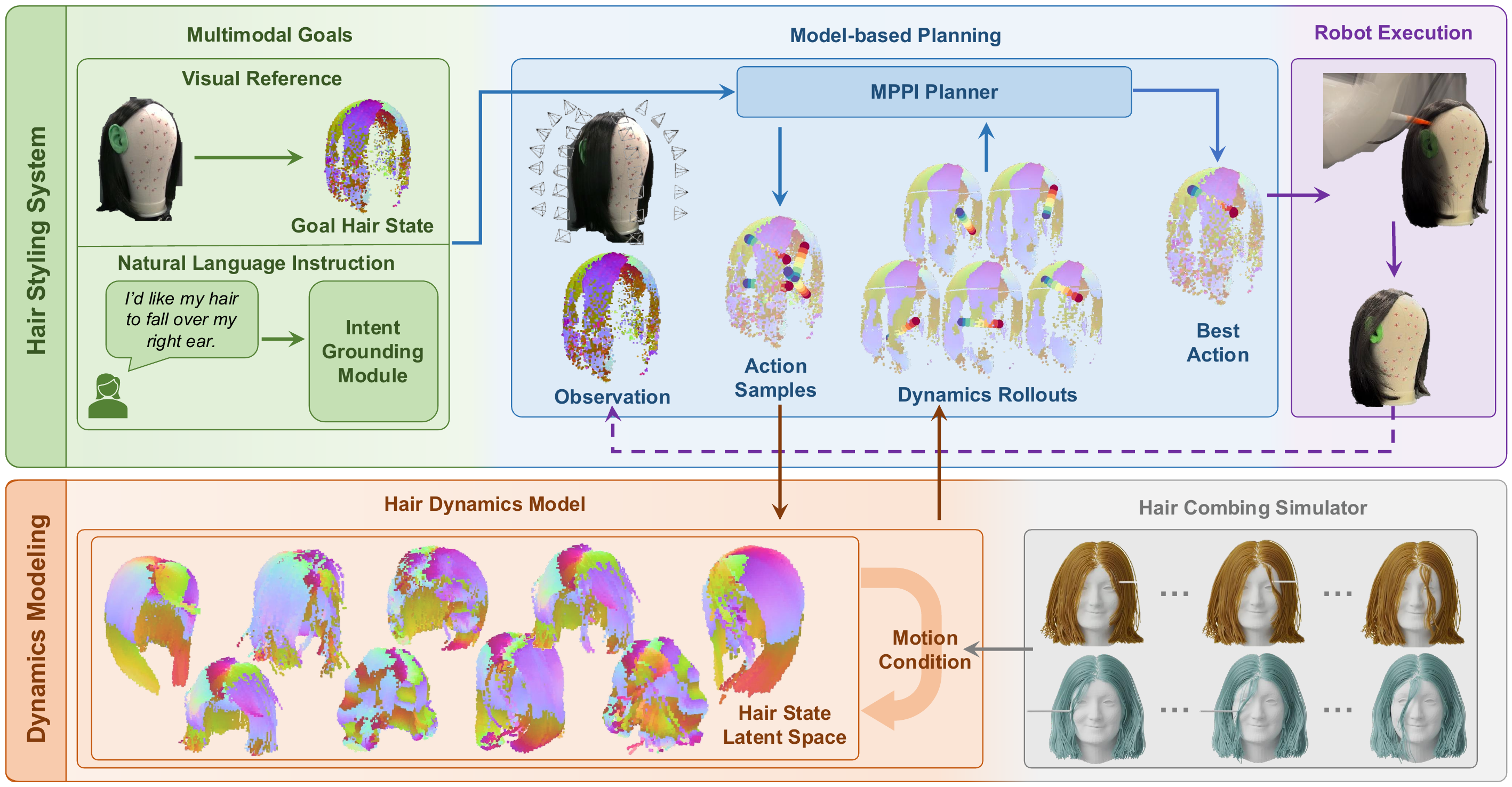}
    \caption{\textbf{Overview.} 
    We introduce \textbf{\systemname}, a modularized, model-based robot hair care system. We propose the first 3D volumetric hair-combing dynamics model, featuring a novel learning approach. It uses an action-conditioned latent state editing mechanism, coupled with a compact 3D latent space over a wide range of hair states, enabled by a hair-combing simulator, for generalizable dynamics modeling. 
 Building on this model, we develop~\systemname{} with a MPPI-based planner for closed-loop multimodal goal-conditioned hair styling, supporting both visual references and natural language instructions as user input.
 }
    \label{fig:teaser}
\end{figure*}

\section{Introduction}
\label{sec:introduction}

Hair is often central to personal identity and self-esteem~\citep{mcfarquhar2000effect, yoo2024inclusion}, yet independently performing routine hair care tasks can be difficult for individuals with limited mobility due to reduced coordination, strength, and flexibility~\citep{waugh2013personal}. To improve accessibility and autonomy, researchers have explored robot hair care systems~\citep{dennler_2021_haircomb, hughes_2021_detanglehair, yoo_2025_moehair, kim_2025_fronthairstyle}, but existing methods rely on either handcrafted trajectories or rule-based controllers, restricting generalization across diverse hairstyles and goals.
These systems also offer limited means for users to flexibly specify their goals, such as through natural language, which constrains their practicality in real-world assistive scenarios.

To address these limitations, we present~\textbf{\systemname}, a model-based robot hair care system. 
The system supports both visual references and natural language instructions, and generalizes across previously unseen hairstyles in simulation and physical wigs varying in color, length, and texture. 
At its core is a learned dynamics model that predicts how hair deforms under different combing motions, enabling the robot to plan actions based on their anticipated effects rather than relying on predefined manipulation rules.

Accurately modeling hair dynamics is challenging because hair consists of thousands of thin, densely interacting strands with complex geometry and partially observable physical properties. Although physics-based simulation can capture detailed strand interactions, its computational cost makes repeated simulation impractical for real-time model-based control. Learning-based neural dynamics models offer a promising alternative~\citep{shi2023robocook, zhang2025particlegrid, tian2025diffusion}, but hair presents several challenges:
1) \emph{Representation.} Sparse point clouds provide limited information about the dense geometry and local strand orientations that characterize hair configurations.
2) \emph{Scalability.} Explicitly modeling the large number of strands and their interactions can be computationally prohibitive, particularly for graph-based approaches requiring dense message passing.
3) \emph{Supervision.} Hair manipulation induces highly localized deformations that can be obscured by global error metrics, while establishing point-wise correspondence across hair states is impractical.
4) \emph{Data.} Hair entanglement and the difficulty of resetting deformed configurations make large-scale real-world data collection costly and time-consuming.

To address these challenges, we introduce a volumetric dynamics learning method that decouples hair-state representation learning from action-conditioned dynamics prediction. We present a 3D hair-combing dynamics model that leverages large-scale synthetic data for dynamics learning and generalizes across various hairstyles. 
We represent hair as a high-resolution volumetric occupancy grid with a 3D orientation field that captures both hair position and local strand flow, which are central to how hair deforms and give a more physically descriptive state than sparse points alone.
This representation design also allows dense supervision on both occupancy and orientation, providing sufficient local signals for the dynamics model to capture fine-grained deformations during learning.
Inspired by ControlNet~\citep{zhang_2023_controlnet}, we pre-train a compact 3D latent space over diverse hair states and introduce a control branch that models dynamics as action-conditioned state editing, enabling generalization across previously unseen hairstyles.

To support large-scale dynamics learning, we develop the first GPU-parallelized simulator for robot hair care, built on the position-based dynamics (PBD) method~\citep{muller2007pbd} within the Genesis framework~\citep{Genesis_2024}. The simulator adapts physics-informed stretch, bending, and twist constraints for strand-level, contact-rich hair manipulation, enabling efficient simulation of complex hair deformations across diverse hairstyles. It provides large-scale synthetic data for dynamics learning and a simulation environment for evaluating closed-loop robot hair styling.

Experiments in simulation show that the proposed model outperforms baselines on generalizable dynamics modeling for local hair deformation across multiple unseen hairstyles.

Building on the proposed dynamics model, we introduce~\systemname, a modularized, model-based robot hair care system for multimodal goal-conditioned hair styling. We adopt a Model Predictive Control (MPC) framework and use a Model Predictive Path Integral (MPPI)-based planner to optimize an action trajectory that minimizes the geometric distance between predicted hair states and the objective~\citep{shi2022robocraft, zhang2024gsdynamics}. 
For visual reference goals, we estimate the target hair state from multi-view observations, and use MPPI to plan directly in the 3D volumetric state space. 
To further support natural language goals, we leverage an off-the-shelf vision-language model (VLM)~\citep{gemini31pro2026} and image generation model~\citep{gemini3proimage2025} to ground semantic instructions into concrete goal images for MPPI planning. This grounding bridges high-level user intent and low-level physical control, making the system practical for users.
Simulation experiments on multiple unseen hairstyles show that \systemname{} achieves superior effectiveness and generalizability for closed-loop visual goal-conditioned hair styling compared to all system baselines, with an average of 22\% lower final geometric error and 42\% higher absolute success rate than the state-of-the-art system. 
An online user study with 58 participants further shows that \systemname{} receives statistically significantly higher ratings than VLM-based system baselines for language goal-conditioned hair styling in simulation, improving both outcome achievement and styling progress consistency.
Real-world experiments show that \systemname{} transfers zero-shot from simulation to 7 physical wigs of varied color, length, and texture, with no real-world model fine-tuning, outperforming baselines on both visual and language goals. Notably, when planned actions fall short under real-world physical complexity, \systemname{} replans from the observed failure and drives the hair styling toward the goal over successive steps.

\newcommand{\extended}{\textbf{[Extended]}}

\textbf{Novelty Statement.}
This work is an extended version of the conference proceedings paper,~\citet{zhao2025dymohair} (DOI: 10.1109/ICRA57385.2026.11696219). We summarize our contributions below, with extensions beyond the conference version marked as \extended:

\begin{itemize}[leftmargin=*,itemsep=0em]

    \item A study of model-based approaches for robot hair manipulation.  

    \item \textbf{\systemname}, a modularized, model-based robot system for goal-conditioned hair styling, supporting:
    
        a) visual goal-conditioned styling from references;
        
        b)~\extended~language goal-conditioned styling, where users specify desired hairstyles through natural language, making the system more practical (\cref{sec:system-language-branch}).

    \item A 3D generalizable volumetric dynamics model for hair combing, with:

        a) baselines as ablations validating latent space pre-training and fine-grained state-action fusion;

        b)~\extended~an ablation on the hierarchical latent structure, showing a more expressive latent space and better dynamics prediction (\cref{sec:ablation-dynamics-hier-structure}).

    \item A GPU-parallelized hair simulator with adapted physics-informed PBD constraints for strand-level, contact-rich hair-combing simulation.

    \item Extensive experiments across multiple hairstyles in simulation and the real world, including:

        a) a simulation-based evaluation of multiple variants of a visual goal-conditioned hair styling system, driven by heuristic rules and different hair dynamics models.

        b)~\extended~a user study to evaluate the language goal-conditioned styling capability, assessing both the robot's executed styling behavior and the quality of intermediate semantic intent grounding (\cref{sec:exp-sim-closed-loop-language});

        c)~\extended~an expanded real-world evaluation to a wider range of hair types varying in color, length, and texture, under both visual and language goals, providing stronger evidence of the zero-shot sim-to-real transferability (\cref{sec:exp-real-world-closed-loop}).

\end{itemize}

\section{Related Works}
\label{sec:related_works}

\subsection{Dynamics Modeling for Deformable Object Manipulation}

Physics-based modeling offers a first-principles dynamics formulation~\citep{tang_2018_linearcoherent, tiburzio2024model}, but is often impractical for deformable objects due to complex internal structures and unobservable material properties, limiting its real-time use in manipulation settings.
Learning-based methods have gained attention as alternatives~\citep{tian2025diffusion}, with graph-based modeling proving effective in capturing spatial relationships in deformable dynamics~\citep{pfaff2021learningmeshbasedsimulationgraph, sanchez2020learning} for various object types like plasticine~\citep{shi2022robocraft, shi2023robocook}, cloth~\citep{lin2022learning}, rope~\citep{zhang2024adaptigraph}, and plush toys~\citep{zhang2024gsdynamics}.
However, these methods often assume easily observable deformations, use low-resolution state representations, and face scalability issues, all of which pose challenges for hair modeling.
Recent works~\citep{zhang2025particlegrid, tian2025diffusion} explore particle-grid hybrid or diffusion-based modeling to improve scalability, but rely on point-wise correspondence for dense supervision, which is impractical for hair.
We propose a new dynamics learning method for hair that preserves geometric and structural features, avoids the scalability limits of graph-based methods and the point-wise correspondence requirement for dense supervision, and enables generalizable modeling.

\subsection{Robot System for Hair Manipulation}
Robot hair manipulation integrates multiple research areas spanning low-level hardware design and high-level algorithm development, and remains underexplored due to its inherent complexity. 
On the hardware side, one line of work~\citep{yoo_2025_moehair} introduces a soft robot manipulator with a force-estimation algorithm to perform predefined hair manipulation skills safely, without exerting excessive force on users. 
On the algorithmic side, prior methods rely largely on 2D observations and hand-designed rules. Some tackle detangling with sensorized brushes using visual and force feedback~\citep{plumbreyes2021combingdoublehelix,hughes_2021_detanglehair}, but operate under constrained settings and specific initial conditions. Another plans combing trajectories aligned with hair flow~\citep{dennler_2021_haircomb}, but lacks goal-driven flexibility. Most recent work, e.g., ~\citet{kim_2025_fronthairstyle}, introduce rule-based, goal-conditioned planning based on 2D orientation differences, but are restricted to particular hairstyles and goals. 
We advance the algorithmic side: rather than operating on 2D observations with hand-designed rules, \systemname{} captures 3D hair states, learns their dynamics, and uses the learned model for both future prediction and closed-loop control, enabling generalizable, multimodal goal-conditioned styling across multiple unseen hairstyles.

\subsection{Hair Dynamics Simulation}

Hair dynamics simulation is challenging due to hair's fine structure, complex inter-strand interactions, and large strand count.
Physics-based methods have been explored for both clump-level and strand-level modeling. Clump-level methods~\citep{koh_2001_hairstrip, wu_2016_hairmesh} represent hair as large bundles, achieving computational efficiency suitable for real-time applications but failing to capture strand-level hair behavior. Strand-based methods, on the other hand, use physically more accurate representations such as mass-spring models~\citep{rosenblum_1991_hairsimulate, selle_2008_massspring} and Kirchhoff rod models~\citep{bergou_2008_discrete, kugelsttadt_2016_pbdcosseraterod}, enabling more accurate modeling of strand-level dynamics, but remain computationally expensive and inefficient.
Recently, learning-based methods have been explored to accelerate simulation~\citep{lyu2020neuralinterp, stuyck2025quaffurerealtimequasistaticneural}, reduce the need for detailed physical modeling~\citep{Wang_2023_neuwigs}, and improve generalizability~\citep{zhang2025hairformertransformerbaseddynamicneural}. While efficient and visually-realistic, they still fall short of achieving strand-level, physically-accurate modeling.
We develop a PBD-based hair-combing simulator tailored for strand-level, contact-rich dynamics, thus achieving efficient, visually realistic, and physically plausible simulation, and supporting synthetic data generation and closed-loop manipulation experiments.

\section{Problem Formulation}
\label{sec:problem}

We develop a model-based robot manipulation system for hair styling using a preset combing tool. Given a user-specified goal $\mathcal{G}$, expressed as either a visual reference or a natural language instruction, the system operates in closed loop. At each timestep $t$, the system receives the current observation $o_t$ from the environment and estimates the underlying hair state $s_t$. The system then selects the next-step action $a_t$ that best advances toward $\mathcal{G}$, guided by a dynamics model $f_{\mathrm{dyn}}$ capturing the hair deformation behavior under combing,~\ie, $s_{t+1} = f_{\mathrm{dyn}}(s_t, a_t)$. 
The action space consists of 3D combing motions, represented as sequences of tool positions and orientations. 
The system executes $a_t$, receives the updated observation $o_{t+1}$, and repeats the state estimation, action selection, and execution until $\mathcal{G}$ is reached.

\section{Hair Combing Dynamics Modeling}
\label{sec:dynamics}

Modeling hair dynamics is challenging because hair consists of dense, thin strands, and combing induces complex, detailed deformations that vary across hairstyles. In this work, we represent hair as a high-resolution volumetric grid, which captures both spatial distribution and local strand flow while remaining efficient to estimate. Building on this representation, we pre-train a compact hierarchical latent space over diverse hair states, and then model dynamics as action-conditioned editing within this space, enabling generalization to unseen hairstyles.

\subsection{State Representation}

\begin{figure}[t!]
    \centering
    \includegraphics[width=1.0\linewidth]{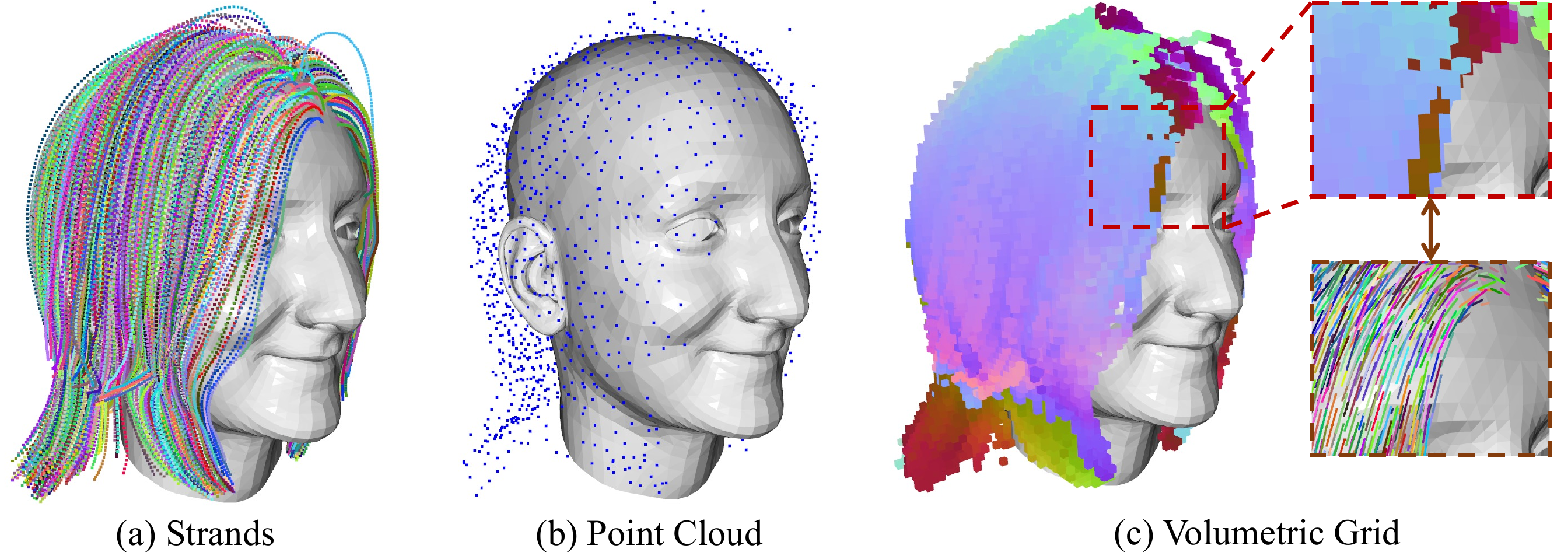}
    \caption{\textbf{Comparison of Hair State Representations.}
    \textbf{(a)} Individual hair strands distinguished by colors.
    \textbf{(b)} A point cloud at 2K-point resolution, the maximum used for point cloud-based methods in our experiments.
    \textbf{(c)} $64 \times 64 \times 128$ grids with about 5\,mm voxel size, where colors denote local strand orientations. Red dashed box: a zoomed-in region. Brown dashed box: the corresponding local strand segments.
    }
    \label{fig:representation}
\end{figure}

Designing an effective state representation is fundamental to a dynamics model: it must be expressive enough to capture task-relevant information, yet reliable and efficient to estimate with a real robot system.
In the context of hair, a strand is the fundamental physical structure from which hair is formed, serving as a key geometric cue for describing hair state. 
However, even the most advanced methods for full 3D strand reconstruction require several minutes~\citep{McGuire2021hairinverse, rosu2022neuralstrands}, making them impractical for a real-time robot manipulation system. 
To achieve both fidelity and efficiency, we draw inspiration from prior computer vision work~\citep{nam_2019_strandaccurate} and represent the hair as a set of dense 3D points, each with a position and a unit direction vector describing the local strand orientation. This form of representation captures both the spatial distribution and the flow direction of hair, and can be estimated robustly and efficiently from multi-view RGB-D data. Compared with widely-used standard point clouds for dynamics modeling, this representation encodes richer geometric and structural information while avoiding the computational cost of strand-level reconstruction.
We further convert the set of points into a high-resolution volumetric grid as the final state representation, following~\citet{saito_2018_hairvae}. Each voxel stores an occupancy value and an orientation:
{
\[
s_t = (occ_t, ori_t), \quad occ_t \in \{0,1\}^{\mathcal{V}_0}, \quad ori_t \in [-1,1]^{\mathcal{V}_0 \times 3},
\]
}%
where $\mathcal{V}_0$ is the spatial resolution of the grid.
This voxelized form allows us to balance computational efficiency with geometric fidelity: the voxels enable the use of highly-optimized neural operators and dense voxel-level supervision in the learning process, while maintaining acceptable accuracy when resolution is sufficiently high. An illustration of the proposed state representation is shown in~\cref{fig:representation}.

\subsection{State Latent Space Pre-training}

\begin{figure}[t!]
    \centering
    \includegraphics[width=\linewidth]{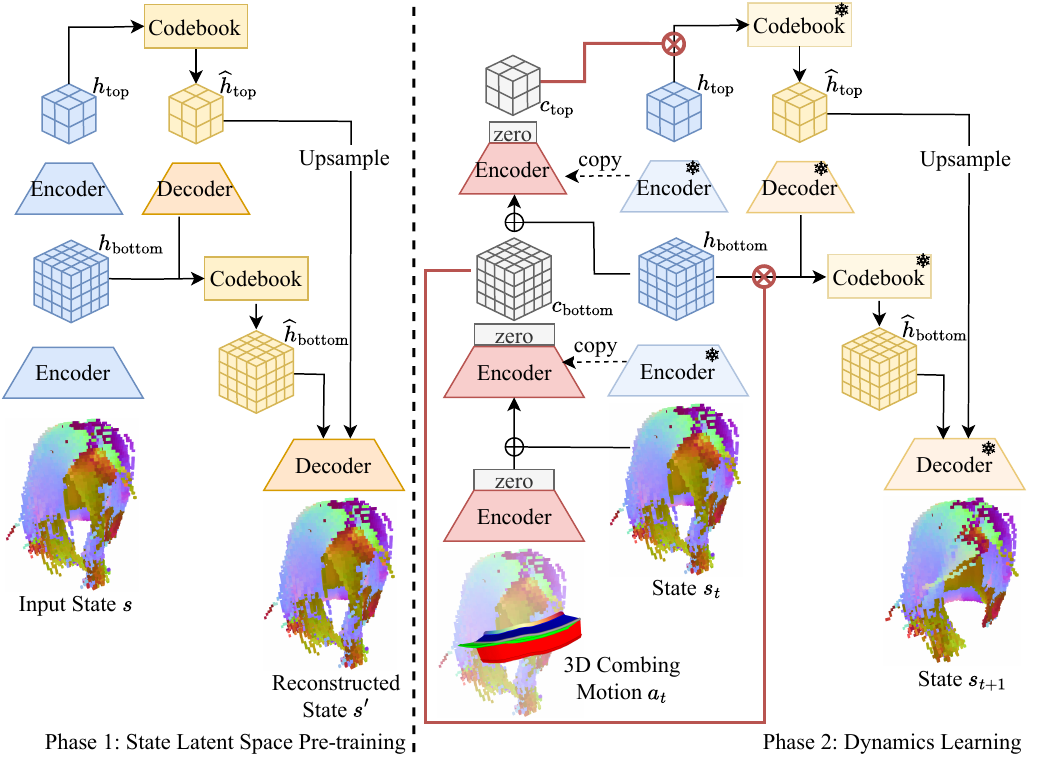}
    \caption{
    \textbf{\systemname's Dynamics Model Overview.} 
\textbf{Left}: State latent space pre-training. A 3D volumetric hierarchical model with vector quantization enables compact compression while preserving detailed representation capability. 
\textbf{Right}: Dynamics learning. The pre-trained model is adapted to capture hair dynamics in a ControlNet-style framework, formulating dynamics as action-conditioned editing in the pre-trained state latent space. \textit{zero}: zero-convolution; \textit{copy}: weight copying for initialization; $\oplus$: element-wise addition; $\otimes$: 3D attention-based feature fusion. In this phase, only the motion encoding path is trainable, with all pre-trained components frozen.
}
\label{fig:dynamics-model}
\end{figure}

Inspired by recent advances in conditioned image and video generation~\citep{zhang_2023_controlnet,geng2024motionprompting,liu2025mosaic}, we propose a novel ControlNet-style~\citep{zhang_2023_controlnet} two-phase method for hair dynamics learning that is compatible with the high-resolution volumetric state representation and enables more generalizable modeling across different hairstyles.
Phase 1 learns a compact latent space of hair states, while Phase 2 learns how robot actions transform states within this space to capture dynamics.

Specifically, in Phase 1, we use pre-training to encode different hairstyles with various deformations during combing into a unified 3D latent space.
In the 3D volumetric setting, we adopt a two-level hierarchical model structure to represent volumetric hair states (see~\cref{fig:dynamics-model}), analogous to VQ-VAE-2~\citep{razavi2019vqvae2} for 2D images. This hierarchy provides compact state compression while preserving both global structure and fine-grained local detail.
The model is pre-trained with state reconstruction as the objective, where, given a volumetric state representation $s$ of any hairstyle with deformation, the model concatenates the occupancy and orientation components together and hierarchically encodes the combined representation into lower resolutions with two 3D encoders, thus producing latent embeddings $h_{\text{bottom}} \in \mathbb{R}^{\mathcal{V}_1 \times D_1}$ and $h_{\text{top}} \in \mathbb{R}^{\mathcal{V}_2 \times D_2}$, where $\mathcal{V}_0 > \mathcal{V}_1 > \mathcal{V}_2$. The higher-resolution bottom embedding captures fine, local structure, while the coarser top embedding captures global structure.
The top-level codebook~\citep{razavi2019vqvae2,van2017vqvae} then quantizes $h_{\text{top}}$ into $\hat{h}_{\text{top}}$ by replacing each entry with its nearest codebook vector, which is further decoded into resolution $\mathcal{V}_1$ to serve as a prior for quantizing $h_{\text{bottom}}$. 
A decoder decodes the quantized $\hat{h}_{\text{bottom}}$ and the up-sampled $\hat{h}_{\text{top}}$ into the final reconstruction $s'$. 
This hierarchical design allows the two quantized latent spaces, \ie, the bottom- and top-level codebooks, to capture complementary local and global information respectively, enabling better state modeling and reconstruction than single-level quantization~\citep{van2017vqvae}.
We use exponential moving averages (EMA)~\citep{van2017vqvae} to update codebooks progressively during training.

\subsection{Dynamics: Action-conditioned State Editing}
\label{sec:dynamics-phase2}

In Phase 2, we formulate the dynamics as an action-conditioned state editing process, and adapt the pre-trained model with a ControlNet-style~\citep{zhang_2023_controlnet} framework to leverage the pre-trained state latent space for capturing hair dynamics.

The pre-trained state encoding path takes the initial state $s_t$ as input and encodes it progressively, as in Phase 1.
For dynamics modeling, we introduce an additional trainable motion encoding path that processes the combing motion $a_t$ as a control signal in parallel with the state, keeping all other pre-trained components, including the codebooks, frozen.

To enforce fine-grained spatial alignment between the state and the motion, we convert the motion into a volumetric grid matching the state resolution before feeding it into the motion encoding path.
Specifically, we sample the motion into up to $K$ uniformly spaced key tool poses and, for each tool pose, compute the shortest distance from every voxel center in the grid to the tool’s center line. Voxels outside a central cylindrical region, defined by the center line and a preset contact radius, are discarded to form a prior for the local contact region. This yields a time-indexed volumetric distance map in $\mathbb{R}^{\mathcal{V}_0 \times K \times 2}$, with the distance value and a binary validity flag marking voxels inside the contact region in the last dimension. 

The motion encoding path consists of three cascaded encoders that process this map: the first preserves resolution, while the other two, sharing the state encoders' architecture, progressively compress the map to lower resolutions, producing control embeddings $c_{\text{bottom}} \in \mathbb{R}^{\mathcal{V}_1 \times D_1}$ and $c_{\text{top}} \in \mathbb{R}^{\mathcal{V}_2 \times D_2}$. 
Following ControlNet~\citep{zhang_2023_controlnet}, the path employs weight copying, zero-convolution, and cross-path feature fusion mechanisms to integrate state and motion features at the fine-grained voxel level while stabilizing early training. The control embeddings are then fused with state embeddings $h_{\text{bottom}}$ and $h_{\text{top}}$ to perform edits in the pre-trained state latent space for the dynamics behavior. The edited embeddings are finally quantized and decoded into the end state $s_{t+1}$.

\subsection{Supervision}

We use the same composite loss for both phases.
For occupancy, we combine the focal loss~\citep{lin2018focallossdenseobject} and the soft Dice loss~\citep{milletari2016vnet} to address class imbalance and encourage volumetric overlap with ground truth, as occupied regions typically form a thin shell within the grid. 
For orientation, we use the L1 loss on normalized orientation vectors for each occupied voxel, with symmetry handling for directional equivalence,~\ie, $v$ and $-v$ represent the same local strand orientation. 
For codebook matching, we adopt the EMA commitment loss from VQ-VAE~\citep{van2017vqvae}, keeping latent entries close to their matched codebook vectors. 
The total loss is a weighted sum of these terms, jointly enforcing geometric accuracy, directional consistency, and latent representation quality.

\section{Hair Dynamics Simulation}
\label{sec:simulation}

Existing neural dynamics models are often trained directly on real-world data for each object. However, for hair, fine-grained strand entanglement and deformation can create messy, hard-to-reset states, making real-world data collection highly time-consuming.
Moreover, the proposed method leverages large-scale data to build an expressive state latent space that broadly covers different hairstyles and various deformations during pre-training, which further increases the impracticality of collecting such data in the real world. 
In this paper, we use fully synthetic data for dynamics learning. We develop a novel GPU-parallelized simulator based on the Genesis framework~\citep{Genesis_2024} to enable efficient strand-level, contact-rich hair-combing simulation, which is both visually realistic and physically plausible.

\begin{figure}[t!]
    \centering
    \includegraphics[width=\linewidth]{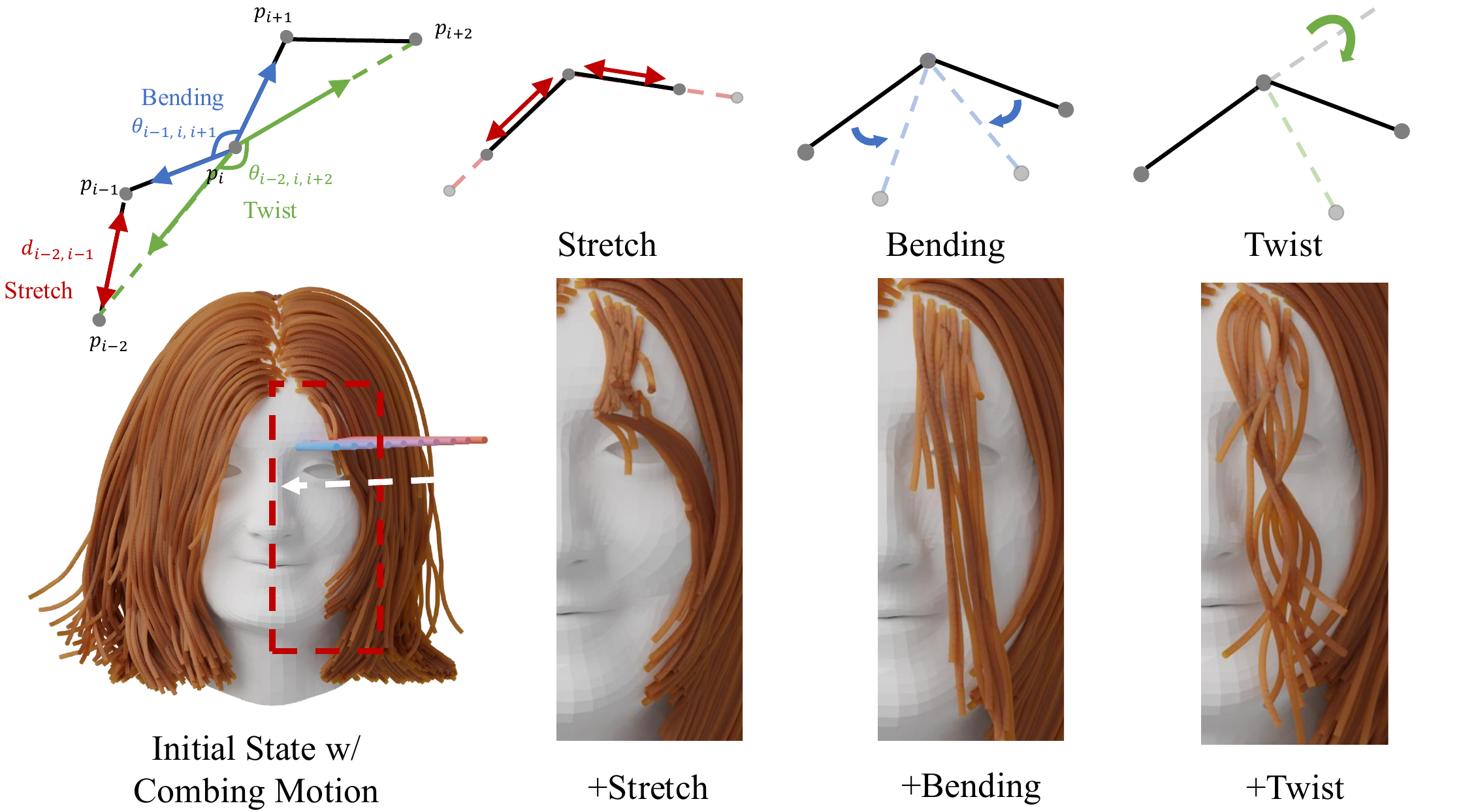}
    \caption{\textbf{Constraints for PBD-based Strand-level, Contact-rich Hair Combing Simulation.}
\textbf{Top:} Each constraint’s formulation and intended effect. 
\textbf{Bottom:} Simulation results under progressive constraint addition. Starting from the initial state (far left), a combing motion is applied along the white dashed arrow (from red to blue). The red dashed box marks the contact-rich region, with its simulation results shown on the right.
With all three constraints, the hair maintains a realistic shape; the twist constraint, in particular, preserves curvature and prevents gravity-induced oversmoothing.
}
    \label{fig:sim-constraint}
\end{figure}

At its core, this simulator builds on the PBD method for strand-level hair simulation.
We model each hair strand as a 3D particle sequence $[p_0, p_1, \cdots, p_n]$, where $p_0$ is the root connected to the scalp and fixed. 
Let 
{\small $d(p_i, p_j) = \|p_i - p_j\|$}
denote the Euclidean distance between particles $p_i$ and $p_j$, and  
{\small $\theta(p_i, p_j, p_k) = \arccos\!\left(\frac{(p_i - p_j)\cdot(p_k - p_j)}{\|p_i - p_j\|\,\|p_k - p_j\|}\right)$}
denote the angle at $p_j$ formed between the vectors $p_i - p_j$ and $p_k - p_j$.  
The corresponding rest-state values are denoted $d^0(p_i, p_j)$ and $\theta^0(p_i, p_j, p_k)$.  
Thousands of strands are simulated in parallel. 
We use three physics-informed constraints to model inner-strand physical properties:

\begin{itemize}[leftmargin=*,itemsep=0em]
   \item \textbf{Stretch:} Neighboring particles preserve their rest distances:%
    {\footnotesize
    \[
    C_{\mathrm{stretch}}(p_i, p_{i+1}) \;=\; d(p_i, p_{i+1}) - d^0(p_i, p_{i+1}), 
    \quad \forall\, i \in [0, n-1].
    \]
    }
        
   \item \textbf{Bending:} Local in-plane bending is regulated by constraining consecutive particle angles:%
    {\footnotesize
    \[
   \begin{aligned}
        C_{\mathrm{bending}}(p_{i-1}, p_i, p_{i+1}) 
        &= \theta(p_{i-1}, p_i, p_{i+1}) \\
        &\quad - \theta^0(p_{i-1}, p_i, p_{i+1}), 
        \quad \forall\, i \in [1, n-1].
    \end{aligned}
    \]
    }
    
   \item \textbf{Twist:} The 3D twist constraint of the strand is approximated by multiple skip-connected 2D bending constraints with a fixed index gap $k>1$:%
    {\footnotesize
    \[
     \begin{aligned}
        C_{\mathrm{twist}}(p_{i-k}, p_i, p_{i+k}) 
        &= \theta(p_{i-k}, p_i, p_{i+k}) \\
        &\quad - \theta^0(p_{i-k}, p_i, p_{i+k}), 
        \quad \forall\, i \in [k, n-k].
    \end{aligned}
    \]
    }  
    They collectively form multiple 3D-intersecting 2D constraint planes to approximate the full 3D twist behavior.
\end{itemize}
The constraints with their effects are illustrated in~\cref{fig:sim-constraint}. 
During simulation, the particle positions are iteratively updated toward satisfying all $C=0$.
Note that the proposed twist model is a heuristic approximation, as true hair twist arises from the strand’s internal physical microstructure. While more complex techniques like Kirchhoff rod models may offer more physically-accurate twist simulation, they typically remain computationally expensive and inefficient. Here, we use the heuristics to balance accuracy and computation efficiency. 
We model inter-strand and hair-tool contacts at the particle level with standard PBD collision and friction handling.

This simulator enables efficient simulation of visually-realistic and physically-plausible hair-combing dynamics across different hairstyles, supplying abundant data for model learning and serving as a testbed for closed-loop experiments with the proposed hair styling system.

\begin{figure*}[t]
    \centering
    \includegraphics[width=\linewidth]{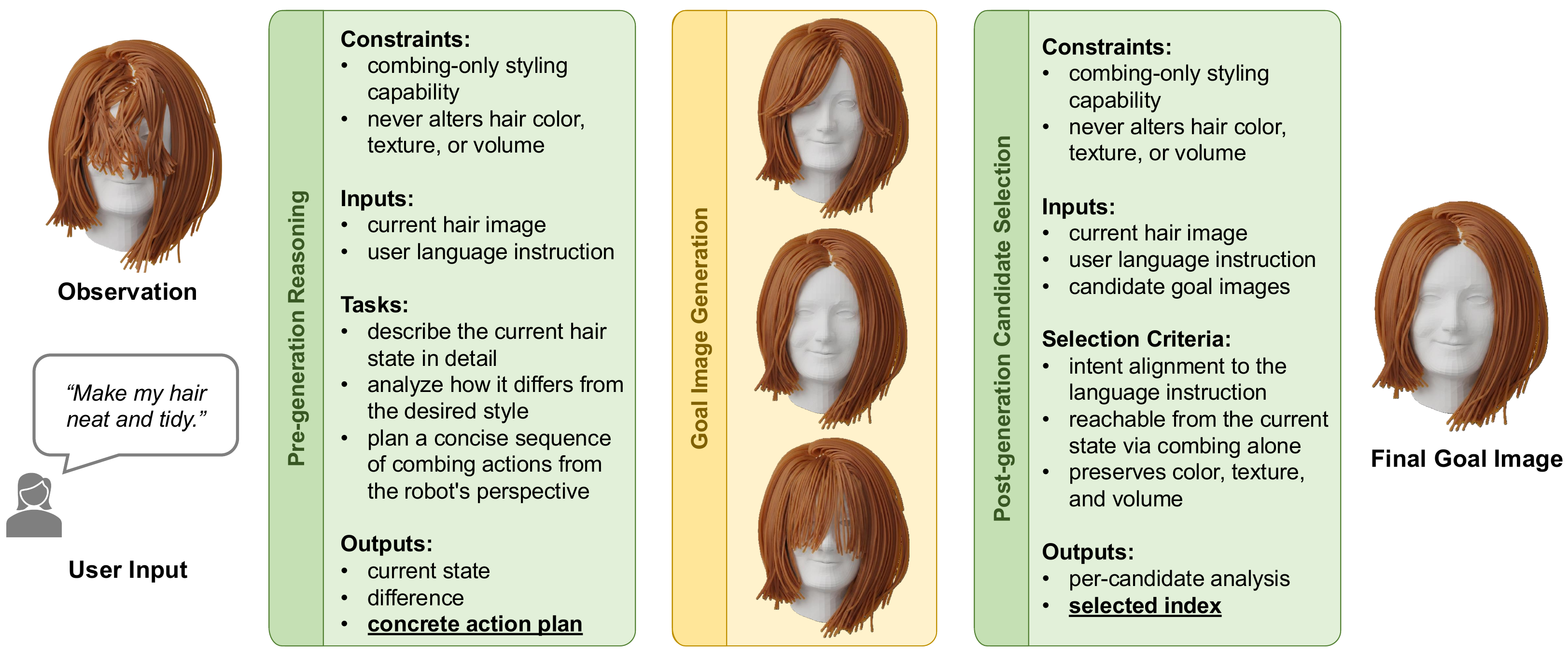}
    \caption{\textbf{Intent Grounding Module of the Language Goal-conditioned Styling Branch.} To translate a high-level language instruction into a physically actionable goal image, we introduce two VLM-based mediation stages around the image generation model.
    \textbf{Mediation stage 1: pre-generation reasoning (left):} the VLM takes the current observation, the user instruction, and the robot's combing-only constraints, and translates the instruction into a concrete sequence of actions.
    \textbf{Goal image generation (middle):} the image generation model is conditioned on this action plan rather than the raw instruction, and queried multiple times for candidate goal images.
    \textbf{Mediation stage 2: post-generation selection (right):} the VLM is reused as a discriminator, analyzing each candidate on intent alignment, reachability via combing alone, and preservation of intrinsic hair properties, then selecting the final goal image, which is handed to the downstream MPPI planner.
    }
    \label{fig:framework-lang-branch}
\end{figure*}

\section{Model-based Robot Hair Styling System}
\label{sec:system}

Building on the proposed hair-combing dynamics model, we develop~\systemname{}, a model-based robot hair care system for hair styling, capable of handling different hairstyles and multimodal goal configurations specified through either visual references or natural language instructions. 
As illustrated in~\cref{fig:teaser}, the system takes multi-view RGB-D observations from the environment and estimates the current volumetric hair state, following~\citet{nam_2019_strandaccurate}. 
Given the current state and a goal configuration, the MPPI-based planner samples candidate actions, rolls out the dynamics model to predict possible future outcomes, and optimizes an action trajectory that minimizes the distance between the predicted states and the goal, following prior neural dynamics-based planners~\citep{shi2022robocraft, zhang2024gsdynamics, zhang2024adaptigraph}. 
The system then executes a chunk of the planned actions, updates its observation, and repeats this loop until the error to the goal falls below a predefined threshold or the action budget is exhausted.

A central design choice in~\systemname{} is to support multimodal goal configurations within a single planning framework. Both visual and language branches translate user input into a spatial goal that the shared MPPI planner consumes; they differ only in the spatial domain where the planning cost is computed. The visual branch operates directly in the 3D volumetric state space of the dynamics model, while the language branch grounds the instruction into a 2D goal image and computes the cost in the image domain. We describe each branch below.

\subsection{Visual Goal-conditioned Hair Styling}
\label{sec:system-visual-branch}
In the visual branch, the system takes a physical hair reference as the goal. We apply the same multi-view RGB-D pipeline used for runtime observation to estimate the goal 3D hair state, which lies in the same volumetric state space as the dynamics model's predictions. MPPI planning therefore operates directly in this space, with a cost defined as the strand-level distance between strand segments reconstructed from the predicted and goal states, jointly incorporating positions and local orientations into a dense planning signal.

\subsection{Language Goal-conditioned Hair Styling}
\label{sec:system-language-branch}
To make the system more practical, we introduce a language goal-conditioned styling branch, in which the user specifies the desired hairstyle through a natural language instruction like ``make my hair neat and tidy.'' 
This offers a convenient alternative to providing a visual reference, but it introduces a new challenge: grounding the high-level semantic intent in language into a physically actionable goal for execution.

The proposed method builds on a key observation: recent image generation models~\citep{gemini3proimage2025} have learned rich semantic relationships between hairstyles and natural language descriptions. We exploit this prior by instantiating the user's instruction as a concrete goal image conditioned on the current observation, then feeding it to the downstream planner. This sidesteps the need to build a task-specific semantic space that is both generalizable across hairstyles and amenable to language-conditioned planning, instead inheriting the semantic priors of large-scale pretrained models.

While these models perform well in general, directly conditioning them on the raw observation and instruction raises two issues. The first is per-sample quality: a generated goal does not always achieve both intent alignment (faithfully reflecting the user's instruction) and physical feasibility (remaining within the robot's combing-based styling capabilities). The second issue is cross-sample variation: publicly available image generation APIs exhibit non-negligible stochasticity, with generated images varying substantially across samples under identical inputs.

We address both with VLM-based mediation before and after image generation, as shown in~\cref{fig:framework-lang-branch}. To improve per-sample quality, we prompt a pretrained VLM (Gemini 3.1 Pro~\citep{gemini31pro2026}) to reason about the user's instruction alongside the current observation and translate the abstract intent into a concrete styling plan, given high-level constraints on the robot's capabilities, namely that it operates by combing alone and cannot alter intrinsic hair properties. This plan, rather than the raw instruction, conditions the image generation model (Gemini 3 Pro Image~\citep{gemini3proimage2025}). To handle cross-sample variation, we sample multiple candidate goal images per instruction and reuse the VLM as a discriminator that selects the candidate most consistent with both the user's intent and the robot's capabilities, yielding the final target for planning.

Since the target now lives in the 2D image domain rather than 3D volumetric space, MPPI planning in this branch uses a correspondingly modified cost. At each iteration, the predicted 3D hair state from the dynamics rollout is projected onto the image plane to produce a predicted hair mask, and the cost is the pixel-level Chamfer Distance between this mask and the goal mask. This preserves the geometric structure of the visual branch's cost while accommodating the 2D nature of the generated goal. 
The language branch thus inherits the semantic priors of pretrained foundation models for high-level intent grounding while relying on the dynamics model for low-level physical planning.

\section{Experiment Setup}
\label{sec:experiment-setup}

\subsection{Overview}

We evaluate~\systemname{} through a comprehensive set of experiments spanning dynamics prediction, closed-loop goal-conditioned manipulation, and zero-shot sim-to-real transfer. Our evaluation examines three key questions: (1) whether the learned dynamics model generalizes to unseen hairstyles and accurately predicts local hair deformations; (2) whether the improved dynamics modeling translates into more effective closed-loop styling toward visual and language goals; and (3) whether the simulation-trained system transfers to physical hair with different appearances and properties.

\begin{figure}[t]
    \centering
    \includegraphics[width=\linewidth]{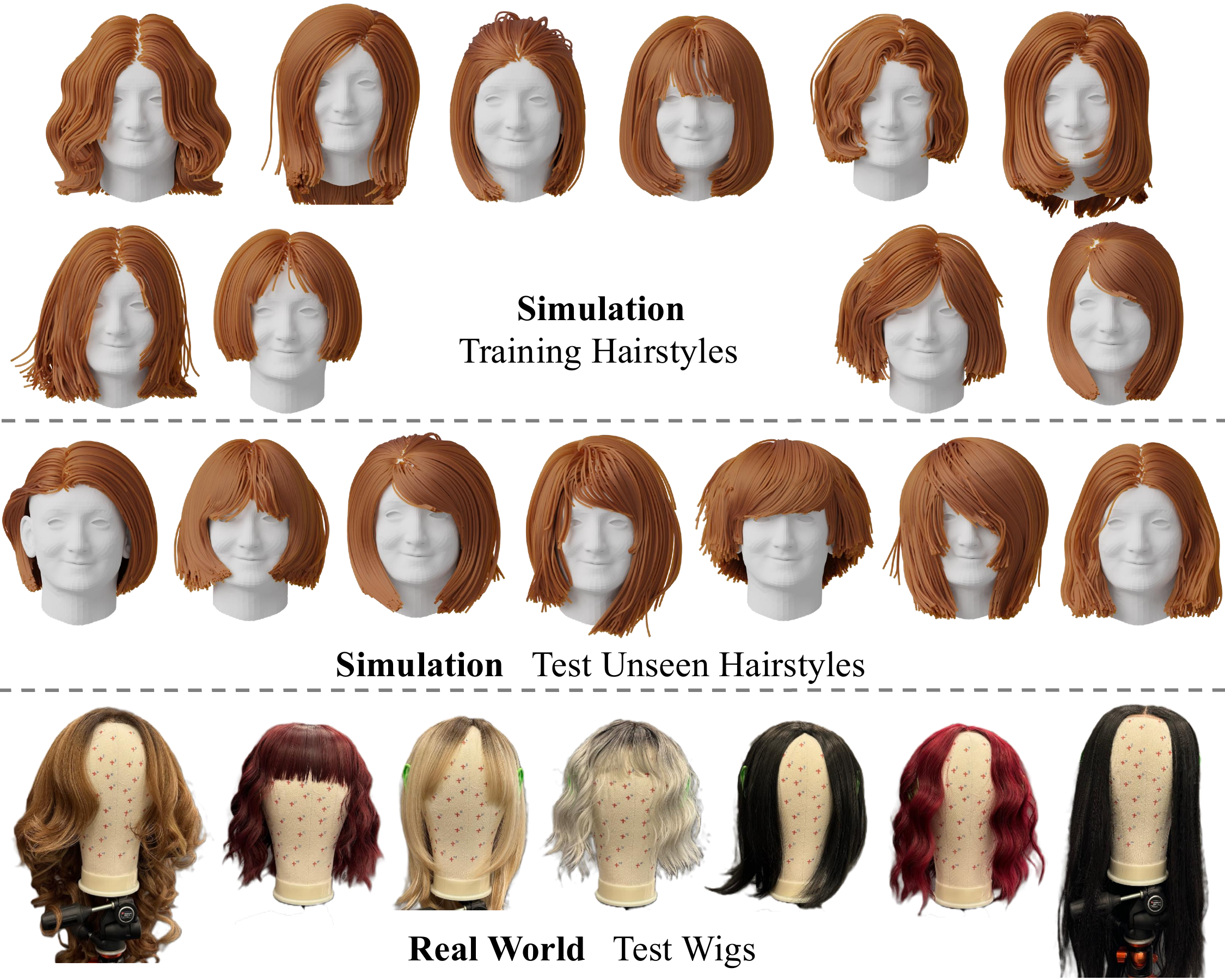}
    \caption{\textbf{Hairstyles Used in Experiments.} \textbf{Top:} 10 synthetic hairstyles used for dynamics model training.
    \textbf{Middle:} 7 unseen synthetic hairstyles used for dynamics model and closed-loop system evaluation in simulation.
    \textbf{Bottom}: 7 physical wigs used for real-world closed-loop system evaluation.
    }
    \label{fig:exp-hair-gallery}
\end{figure}

For dynamics learning, we train the model on 10 synthetic hairstyles and evaluate on 7 unseen hairstyles, comparing it against three dynamics baselines and conducting architectural ablations. For visual goal-conditioned styling, we evaluate five system variants on 7 unseen hairstyles, with 3 initial conditions and 5 random seeds per hairstyle, totaling 525 closed-loop simulation trials. For language goal-conditioned styling, we evaluate 144 demonstrations across 8 tasks and 3 random seeds, complemented by two 58-participant user studies assessing both styling outcomes and semantic intent grounding. Finally, we evaluate the zero-shot sim-to-real transfer on 7 physical wigs varying in color, length, and texture, under both visual and language goals, including comparisons against rule-based and VLM-based systems. Together, these experiments assess the proposed method at the levels of dynamics prediction, closed-loop task performance, and real-world deployment. \cref{fig:exp-hair-gallery} shows the hairstyles used in our experiments.

\subsection{Simulation Setup}
\label{sec:exp-setup-sim}

The simulation workspace consists of a mannequin head and a hair model with a realistic hairstyle, as shown in~\cref{fig:exp-setup}. 
The tool for manipulation is a thin cylinder. 
All hair models and the mannequin head are from the USC-HairSalon database~\citep{hu2015single}.
For simulation stability, we preprocess each strand for an appropriate linear density while preserving its curvature.

\subsection{Real-world Setup}
\label{sec:exp-setup-real}

\begin{figure}[t]
    \centering
    \includegraphics[width=\linewidth]{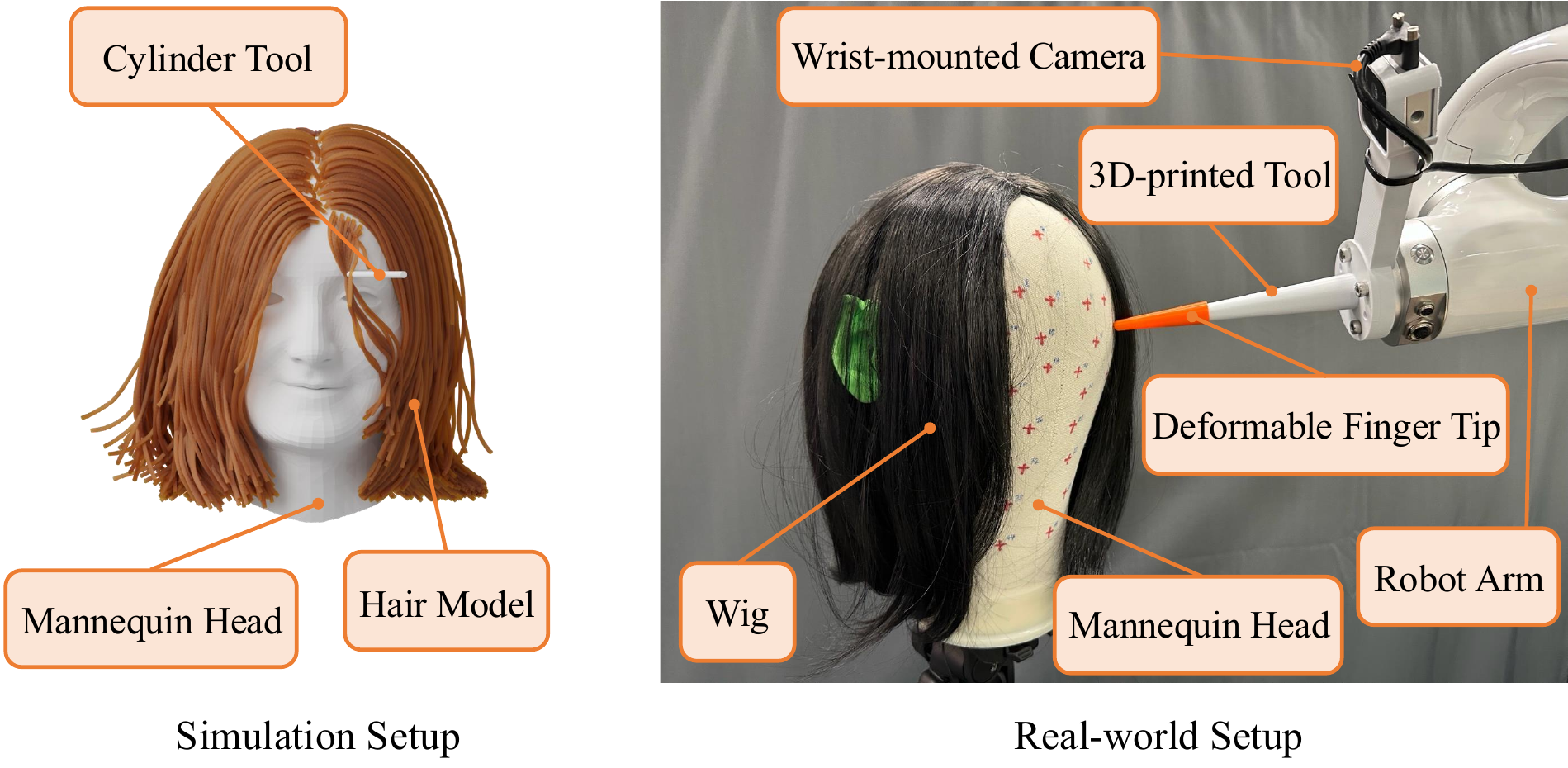}
    \caption{\textbf{Experiment Setup.} \textbf{Left}: simulation setup. \textbf{Right}: real-world setup.
    }
    \label{fig:exp-setup}
\end{figure}

\cref{fig:exp-setup} also shows our real-world workspace, consisting of a 25\,cm$\times$19\,cm$\times$30\,cm mannequin head with a physical wig. The head is painted with calibration marks.
We use a fixed UFactory 850 6-axis robot arm with a 3D-printed tool of length 16\,cm and tip radius 4\,mm, designed to resemble the cylinder tool used in simulation to reduce the sim-to-real gap. We add a TPU-printed deformable finger tip on it to enable better contact between the tool and the head. 
For perception, we use a wrist-mounted RealSense D405 RGB-D camera to capture multi-view observations of the hair by varying the robot pose. 
Both the head and the camera are calibrated to the robot frame.

\section{Experiments on Dynamics Learning}
\label{sec:exp-dynamics}

\subsection{Data}
\label{sec:exp-dynamics-data}

We construct a large-scale synthetic dynamics dataset using the hair-combing simulator (\cref{sec:simulation}) for generalizable dynamics learning. Specifically, for each of 10 training hairstyles, we randomly sample over 1K diverse tool motions, rolling them out in simulation to alter the hair state, and recording intermediate states and sub-motions. After filtering low-quality samples, we obtain a dynamics dataset of 10K transitions.
For Phase 1 latent space pre-training, we combine the 10 undeformed training hairstyles, their deformed variants from the simulated transitions, and an existing large-scale rest-state hairstyle database~\citep{he2025perm,Hair20k}, yielding 47K hair states to build an expressive latent space. We then train the Phase 2 dynamics model on the 10K-transition dataset above.
For evaluation, we generate a separate dynamics dataset from 7 unseen hairstyles, producing 800 filtered transitions to test the model's generalizability.

\subsection{Baselines}
\label{sec:exp-dynamics-baseline}

We compare the proposed method against three baselines:
\begin{itemize}[leftmargin=*,itemsep=0em]
    \item \textbf{PC-GNN}: represents hair as point clouds with per-point orientation, applies a GNN-based structure without pre-training, following the most common method for neural deformable object dynamics~\citep{shi2022robocraft}.
    \item \textbf{V-UNet}: uses the same volumetric hair representation as the proposed method, but with a UNet-based architecture, pyramid fine-grained state-action fusion, and no pre-training; ablates the benefit of the pre-trained latent space.
    \item \textbf{V-FiLM}: uses the same volumetric representation and pre-trained latent space as the proposed method, but with a FiLM~\citep{perez2018film}-style state-action fusion mechanism; ablates the effectiveness of the ControlNet-style fusion design for fine-grained state-action fusion.
\end{itemize}

\subsection{Evaluation Metrics}
We convert the outputs of all volumetric methods into point clouds with per-point orientation and down-sample them to the same resolution as \textbf{PC-GNN} for fair comparison. Three metrics are used: 
1) $\mathrm{CD}_\mathrm{point}$: Chamfer Distance (CD) between predicted and ground-truth point clouds, ignoring orientation.
2) $\mathrm{Err}_\mathrm{ori}$: point-level orientation error, computed as the average angular difference between predicted orientations and those of the closest ground-truth points, with symmetry tolerated. Predicted points farther than 2\,cm from the ground truth are discarded to avoid invalid matches.
3) $\mathrm{CD}_\mathrm{strand}$: strand-level CD, where predicted point clouds with orientations are reconstructed into strand segments to evaluate the strand structure they convey, measuring on average how well each predicted segment aligns with its closest ground-truth segment and vice versa.  

\subsection{Implementation Details}  
For \textbf{PC-GNN}, we use 2K-point resolution, the highest feasible while maintaining enough message-passing capability under the same computation budget as other methods. For all volumetric methods, we use $64 \times 64 \times 128$ grids with a $\sim$5\,mm voxel size. All models are trained to convergence on 2 NVIDIA RTX 4090 GPUs.

\begin{table}[t]
\footnotesize
\centering
\caption{\textbf{Dynamics Model Evaluation on Unseen Hairstyles}}
\label{table:dynamics-quant}
{\renewcommand{\arraystretch}{1.2}
\centering
\resizebox{0.48\textwidth}{!}{
\begin{tabular}
{c@{\hspace{10pt}}c@{\hspace{10pt}}c@{\hspace{10pt}}c@{\hspace{10pt}}c@{\hspace{10pt}}c@{\hspace{10pt}}c}
\shline
\multirow{2}{*}{Method} &
\multicolumn{2}{c}{$\mathrm{CD}_{\mathrm{point}} \downarrow$} &
\multicolumn{2}{c}{$\mathrm{Err}_{\mathrm{ori}} \downarrow$} &
\multicolumn{2}{c}{$\mathrm{CD}_{\mathrm{strand}} \downarrow$} \\
& mean & 90th & mean & 90th & mean & 90th \\
\hline

PC-GNN    & 0.0814 & 0.1359 & 13.34 & 30.16 & 0.1052 & 0.1983 \\
V-UNet    & 0.0792 & 0.1345 & 14.25 & 31.00 & 0.1047 & 0.1966 \\
V-FiLM    & 0.0807 & 0.1334 & 13.60 & 29.59 & 0.1065 & 0.2011 \\
Proposed      & \textbf{0.0775} & \textbf{0.1240} & \textbf{12.03} & \textbf{26.12} & \textbf{0.1005} & \textbf{0.1878} \\

\shline

\end{tabular}
}}
\end{table}

\subsection{Results}
\label{sec:exp-dynamics-results}
\begin{figure}[t!]
    \centering
    \includegraphics[width=\linewidth]{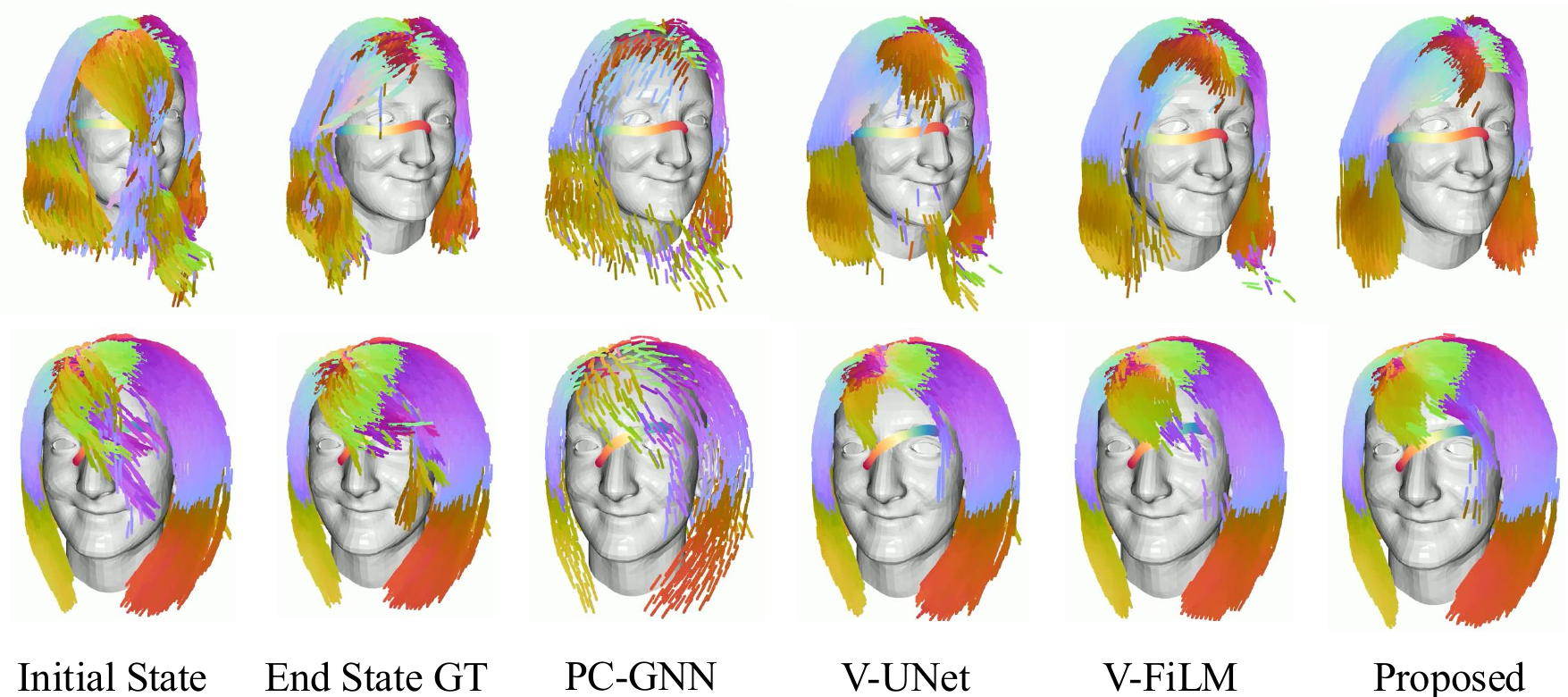}
    \caption{
    \textbf{Qualitative Evaluation of Dynamics Learning on Unseen Hairstyles.} 
From left to right: the initial state, the ground-truth end state, and predicted end states from different methods.
Local strand segments are color-mapped by orientation; due to discontinuities in the mapping, strands with similar orientations may occasionally appear in different colors. 
As a result, local orientation is best assessed from the strand segment geometry rather than the color map.
}
\label{fig:dynamics-qual-result}
\end{figure}

We evaluate all methods on 7 unseen hairstyles to test generalizability, with results shown in~\cref{table:dynamics-quant}. 
Since in hair-combing scenarios deformations are highly localized, occurring mainly near the comb while most of the hair remains stable and unchanged, evaluating the entire hairstyle can obscure local effects.
We therefore focus on the near-motion region and report both the mean, reflecting overall performance, and the 90th percentile, which captures sparse but significant deformations, e.g., deformations that occur in only a few strands, and reduces averaging bias.
The results show that the proposed method outperforms all baselines in capturing local hair deformation behavior in the near-motion region for unseen hairstyles. 
The proposed method surpasses the widely used \textbf{PC-GNN} method, underscoring its suitability for generalizable hair dynamics. 
Compared with other voxel-based methods, it leverages latent space pre-training for stronger generalization than \textbf{V-UNet}, and finer state-action fusion for more accurate future state prediction than \textbf{V-FiLM} based on the same pre-trained state latent space.
We provide a qualitative comparison across all methods in~\cref{fig:dynamics-qual-result}.
Overall, these results highlight the effectiveness and generalizability of the proposed method on hair-combing dynamics modeling, validating the design choices of pre-training and fine-grained state-action fusion.

\subsection{Ablation on Hierarchical Model Structure}

\label{sec:ablation-dynamics-hier-structure}

\begin{figure}[t]
    \centering
    \includegraphics[width=\linewidth]{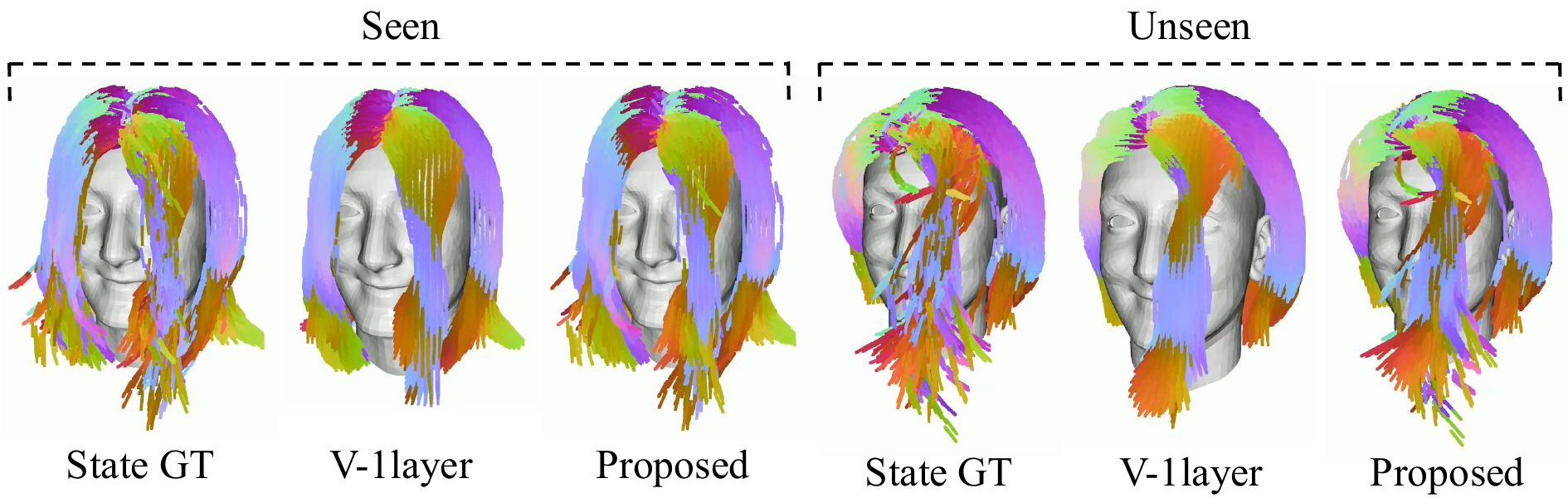}
    \caption{
    \textbf{Qualitative Evaluation of the Hierarchical Model Structure Ablation in Dynamics Learning.} 
For both seen and unseen hairstyles, we show the ground-truth state alongside reconstructions from \textbf{V-1layer} and the proposed method.
}
\label{fig:ablation-hier-dynamics-qual-result}
\end{figure}

We further ablate the hierarchical model structure to evaluate its benefit. 
We introduce \textbf{V-1layer}, which is identical to the proposed method in all other aspects (\eg, pre-training and the ControlNet-style framework) but replaces the hierarchical two-layer structure with a single-layer architecture. 
Specifically, \textbf{V-1layer} removes the bottom-level codebook and instead encodes the initial state directly into a top-level embedding, quantizes this embedding with a single codebook whose size equals the combined size of the top- and bottom-level codebooks within the proposed method, and decodes the result for prediction.
Matching the total codebook capacity in this way ensures the comparison isolates the effect of the coarse-to-fine hierarchical factorization rather than differences in representational capacity.
We compare \textbf{V-1layer} against the proposed method on both latent space pre-training quality and dynamics learning performance. 
For pre-training, we assess reconstruction on 1,000 cases with diverse deformations, covering 10 seen and 7 unseen hairstyles, at the same 2K-point resolution used previously; higher reconstruction accuracy indicates a more expressive latent space. 
For dynamics learning, we evaluate end-state prediction under the same protocol as before.
As shown in~\cref{table:dynamics-ablation-hierarchy}, the hierarchical structure yields a stronger latent space, achieving higher reconstruction accuracy on both seen and unseen hairstyles and in turn improving dynamics learning performance.
We visualize reconstruction examples for seen and unseen hairstyles in~\cref{fig:ablation-hier-dynamics-qual-result}, showing that the pre-trained latent space generalizes well to unseen hairstyles under deformation. 
Overall, compared with the single-layer variant, the hierarchical design builds a richer latent space that preserves finer geometric detail, providing a stronger foundation for downstream dynamics learning.

\begin{table}[t]
\footnotesize
\centering
\caption{\textbf{Ablation on Hierarchical Model Structure in Dynamics Learning}}
\label{table:dynamics-ablation-hierarchy}
{\renewcommand{\arraystretch}{1.2}
\centering
\resizebox{0.48\textwidth}{!}{
\begin{tabular}
{c@{\hspace{10pt}}c@{\hspace{10pt}}c@{\hspace{10pt}}c@{\hspace{10pt}}c@{\hspace{10pt}}c@{\hspace{10pt}}c@{\hspace{10pt}}c}
\shline
\multirow{2}{*}{Split} & \multirow{2}{*}{Method} &
\multicolumn{2}{c}{$\mathrm{CD}_{\mathrm{point}} \downarrow$} &
\multicolumn{2}{c}{$\mathrm{Err}_{\mathrm{ori}} \downarrow$} &
\multicolumn{2}{c}{$\mathrm{CD}_{\mathrm{strand}} \downarrow$} \\
& & mean & 90th & mean & 90th & mean & 90th \\
\shline

\multicolumn{8}{c}{Latent Space Pre-training - Reconstruction} \\ 
\hline

\multirow{2}{*}{Seen} 
& V-1layer    & 0.0694 & 0.1015 & 8.33 & 18.11 & 0.0873 & 0.1485 \\
& Proposed      & \textbf{0.0612} & \textbf{0.0959} & \textbf{6.83} & \textbf{14.79} & \textbf{0.0844} & \textbf{0.1467} \\
\hline

\multirow{2}{*}{Unseen} 
& V-1layer    & 0.0683 & 0.0992 & 9.51 & 19.61 & 0.0824 & 0.1398 \\
& Proposed      & \textbf{0.0598} & \textbf{0.0936} & \textbf{7.49} & \textbf{15.89} & \textbf{0.0787} & \textbf{0.1362} \\
\shline

\multicolumn{8}{c}{Dynamics Learning - End State Prediction} \\ 
\hline

\multirow{2}{*}{Seen} 
& V-1layer    & 0.0791 & 0.1273 & 9.75 & \textbf{20.54} & 0.0981 & 0.1808 \\
& Proposed      & \textbf{0.0750} & \textbf{0.1183} & \textbf{9.55} & 20.63 & \textbf{0.0975} & \textbf{0.1802} \\
\hline

\multirow{2}{*}{Unseen} 
& V-1layer    & 0.0818 & 0.1371 & 12.84 & 27.06 & 0.1022 & 0.1909 \\
& Proposed      & \textbf{0.0775} & \textbf{0.1240} & \textbf{12.03} & \textbf{26.12} & \textbf{0.1005} & \textbf{0.1878} \\

\shline

\end{tabular}
}}
\end{table}

\section{Experiments on Closed-loop Hair Styling in Simulation}
\label{sec:exp-sim-closed-loop}

\subsection{Visual Goal-conditioned Hair Styling}
\label{sec:exp-sim-closed-loop-visual}

\subsubsection{Baselines.}
We compare~\systemname{} against four baselines.
\textbf{Rule-based System}: the only existing method for visual goal-conditioned robot hair styling~\citep{kim_2025_fronthairstyle}, which uses a single front-view 2D observation and handcrafted rules to derive actions from 2D orientation differences between current and goal states.
\textbf{PC-GNN System},~\textbf{V-UNet System}, and~\textbf{V-FiLM System}: three model-based systems that replace~\systemname{}’s dynamics model with the baseline models described in~\cref{sec:exp-dynamics-baseline}, while keeping all other parts unchanged.

\subsubsection{Implementation Details.}
All systems for visual goal-conditioned hair styling assume geometrically grounded goal configurations that are well-aligned with the states, consistent with existing neural dynamics-based deformable object manipulation methods~\citep{shi2022robocraft, zhang2024gsdynamics, zhang2024adaptigraph}. For all model-based systems, the MPPI planning cost is defined as the strand-level distance between strand segments reconstructed from the predicted future states and the goal state, incorporating both point and orientation predictions in a unified manner.

\begin{figure}[t]
    \centering
    \includegraphics[width=\linewidth]{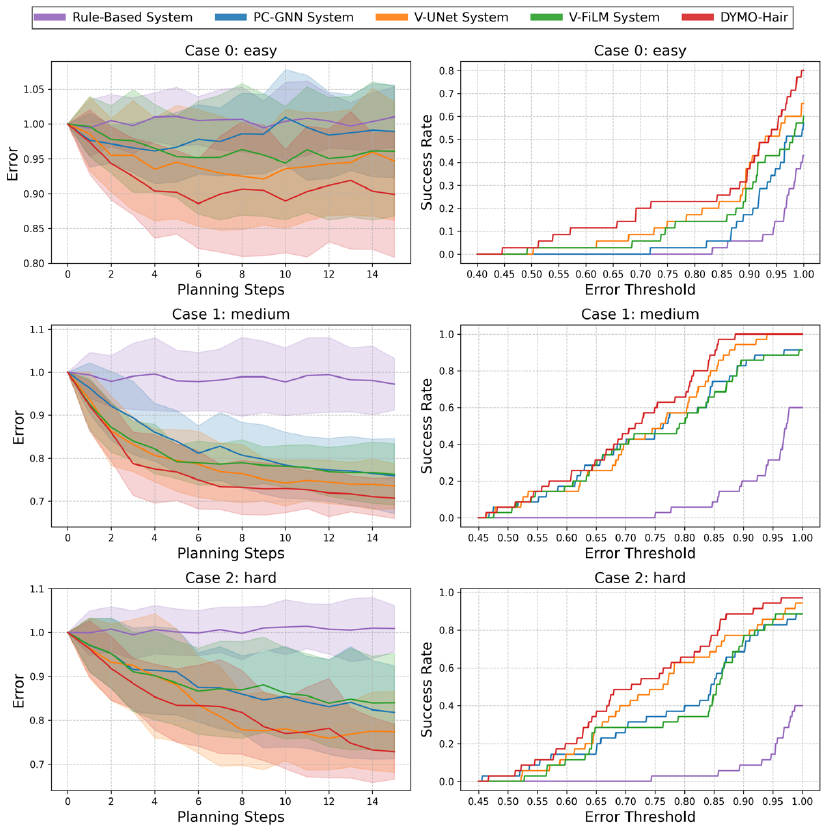}
    \caption{\textbf{Quantitative Results for Visual Goal-conditioned Hair Styling in Simulation.} Experiments are conducted on 7 unseen hairstyles, each with 3 cases repeated 5 times. 
    \textbf{Left:} Error curves over planning steps; solid lines and shaded regions denote the mean and standard deviation across hairstyles, and faster error reduction with lower final error indicates higher effectiveness.
    \textbf{Right:} Success rate curves w.r.t. error thresholds used to determine success, reflecting the distribution of final errors; curves closer to the top-left correspond to more low-error outcomes and thus better performance. 
    In both panels, error is defined as relative error: the ratio of the current strand-level distance to the initial strand-level distance, which provides a unified metric across hairstyles with varying absolute error magnitudes.
    }
    \label{fig:exp-sim-quant-closed-loop-visual}
\end{figure}

\begin{figure*}[t!]
    \centering
    \includegraphics[width=\linewidth]{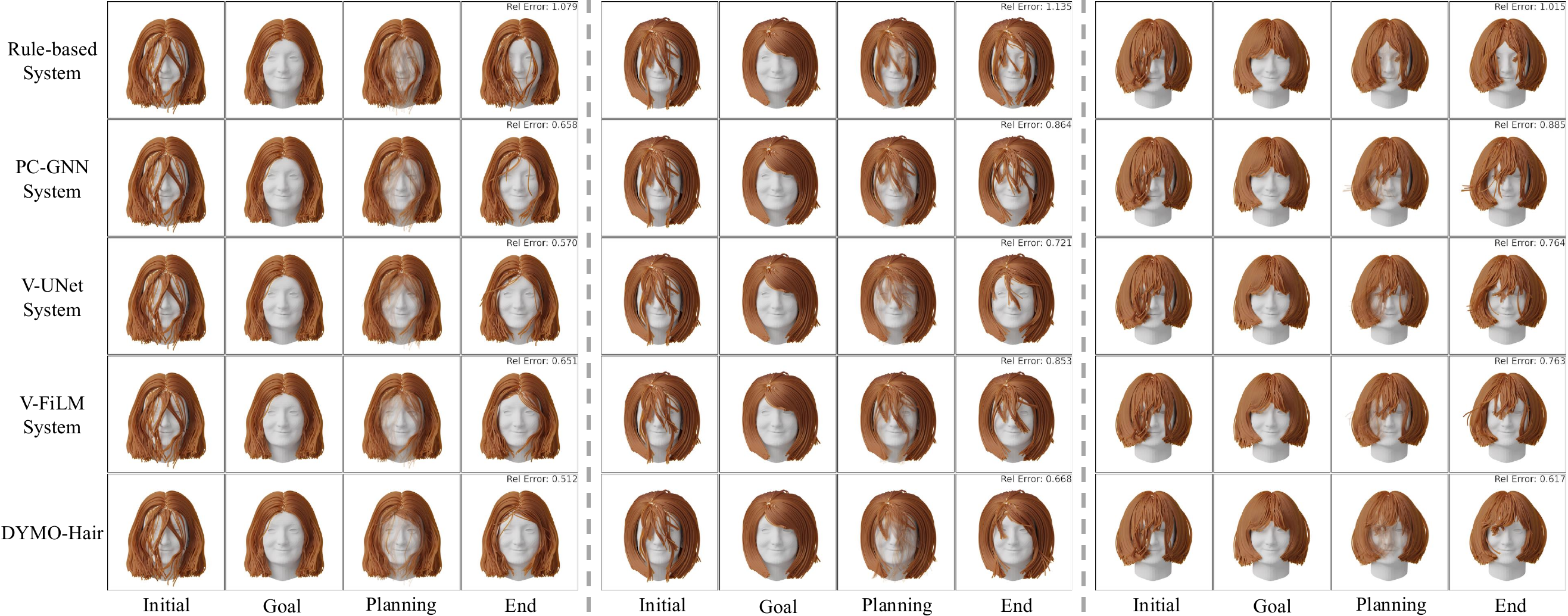}
    \caption{\textbf{Qualitative Results for Visual Goal-conditioned Hair Styling in Simulation.} 
    We show three \textit{hard} cases on different unseen hairstyles. For each, columns from left to right show the initial state, the goal state, a snapshot of the intermediate planning steps, and the styled end state (annotated with its final relative error).
    }
    \label{fig:exp-sim-qual-closed-loop-visual}
\end{figure*}

\subsubsection{Results.} 
We first evaluate~\systemname{}’s effectiveness and generalizability for visual goal-conditioned closed-loop hair styling thoroughly in simulation, using multiple unseen hairstyles and varying initial states. Specifically, we test on 7 unseen hairstyles with 3 cases per hairstyle at different levels of messiness, and repeat each case 5 times with different random seeds to reduce randomness. All experiments are conducted with a maximum budget of 15 steps. 
Quantitative results in~\cref{fig:exp-sim-quant-closed-loop-visual} show that all model-based methods outperform \textbf{Rule-based System}. Among them, \textbf{\systemname{}} achieves the best performance, with an average of 22\% lower final geometric error and 42\% higher absolute success rate across three cases (using thresholds of 0.90, 0.70, and 0.70, chosen as reasonable success criteria based on experiments and visualization), demonstrating both the advantage of incorporating a dynamics model for improving system capability and generalizability, and the superior effectiveness of the dynamics model from the proposed method in boosting closed-loop manipulation at the system level.
Qualitative results are shown in~\cref{fig:exp-sim-qual-closed-loop-visual}.

\subsection{Language Goal-conditioned Hair Styling}
\label{sec:exp-sim-closed-loop-language}

Evaluating language goal-conditioned hair styling is challenging because natural language instructions express complex semantic goals that are difficult to capture with scalar metrics, and a single instruction may admit multiple valid hairstyles rather than a unique ground truth. These challenges make reference-based or automated evaluation difficult, motivating user studies based on human judgments of styling performance.

We organize the evaluation into two studies.
\textbf{Study 1} evaluates \textit{closed-loop styling}: the full system's behavior in styling the hair toward the instruction, from initial observation through planning and execution.
\textbf{Study 2} further evaluates the intermediate \textit{semantic intent grounding}: the quality of the goal image produced by the intent grounding module, which serves as the target for downstream planning and execution. 
Both studies draw on a shared participant pool and a common analysis protocol.

\subsubsection{Participants.}
We recruited respondents via Amazon Mechanical Turk, requiring a $\geq$99\% HIT approval rate, $\geq$1000 approved HITs, self-reported English fluency, and no self-reported visual impairments. After excluding 17 respondents flagged as likely bots (low reCAPTCHA score) or as not engaging meaningfully with the task (extreme completion times), 58 participants (aged 27--65; 37 women, 20 men, 1 preferred not to say) entered the analysis.
The studies were approved by the Carnegie Mellon University Institutional Review Board (IRB Study ID: STUDY2026\_00000069).

\subsubsection{Procedure.}
Prior to the studies, we collected an offline database of styling demonstrations in simulation, fully factorial over methods, tasks, and random seeds. The 8 tasks pair 4 synthetic hairstyles with 2 language instructions each, and every method-task combination was repeated under 3 random seeds to account for stochasticity in goal image generation and planning, yielding $6 \times 8 \times 3 = 144$ demonstrations in total.
Each participant completed both studies in a fixed order. In each study, participants viewed several demonstrations (4 for \textbf{Study 1}, 2 for \textbf{Study 2}), presented one at a time and sampled uniformly from the relevant subset of the database, and rated each on the study-specific constructs, every construct measured by three 7-point Likert items. Further procedural details are provided in the appendix.

\subsubsection{Analysis Protocol.}
Our primary analysis uses linear mixed-effects (LME) models,
which account for the fact that ratings are not independent:
each participant rates several demonstrations, and each demonstration belongs to one of a limited set of tasks. For each construct, we average its three Likert items into a single score and model this as a function of method, with random intercepts for participant, task, and demonstration instance. 
We evaluate planned pairwise comparisons between methods, with Holm correction within each study. For each comparison, we report both the difference in mean rating between the two methods in Likert points ($\Delta$) and its statistical significance.
The full protocol and complete results are in the appendix.

\subsubsection{Study 1: Closed-loop Styling.}
\label{sec:user-study-styling}
This study assesses how well the full system styles hair toward the user's intent through closed-loop execution. We compare \systemname{} against model-free and lower-fidelity model-based variants, isolating the contributions of model-based predictive planning and \systemname{}'s high-fidelity volumetric dynamics model. We find that \systemname{} significantly outperforms both VLM-based variants in both styling outcome and progress toward the goal, while the gain from dynamics-model fidelity is positive but smaller, indicating that predictive planning is the dominant factor.

\paragraph{Constructs.}
We measure two constructs in \textbf{Study 1}, each rated on three 7-point Likert items:
\begin{itemize}[leftmargin=*,itemsep=0em]
\item \textit{Outcome achievement}: how well the final hairstyle reflects the user's intent, judged from the trajectory's final frame.

\item \textit{Progress consistency}: how coherently the system moves toward the intended hairstyle throughout execution, judged from key intermediate frames of the styling trajectory.
\end{itemize}

\paragraph{Baselines.}
We compare \systemname{} against three baselines that probe two axes of the design: how actions are chosen (model-based predictive planning vs.\ VLM-based selection) and the fidelity of the dynamics model for model-based predictive planning (\systemname{}'s volumetric model vs.\ the lower-fidelity PC-GNN).

\begin{itemize}[leftmargin=*,itemsep=0em]
\item \textbf{VLM Planner}: replaces the MPPI planner with a pretrained VLM (Gemini 3.1 Pro~\citep{gemini31pro2026}) that takes the current observation, the language instruction, and the grounded goal image, and selects an action from a set of sampled candidates via PIVOT~\citep{nasiriany2024pivot}-style visual prompting.

\item \textbf{End-to-End VLM}: further removes the intent grounding module, so the same VLM selects actions based only on the current observation and the language instruction, without a grounded goal image.

\item \textbf{PC-GNN Model}: replaces \systemname{}'s volumetric dynamics model with the PC-GNN baseline from~\cref{sec:exp-dynamics-baseline}, holding all other components fixed to isolate the contribution of dynamics-model fidelity for planning.
\end{itemize}

\paragraph{Hypotheses.}
We test \systemname{}'s design along the two axes above: model-based predictive planning versus VLM-based action selection (\textbf{H1}, \textbf{H2}), and the contribution of dynamics-model fidelity (\textbf{H3}). Each hypothesis predicts higher styling quality, measured by the two constructs.

\textbf{H1:} Predictive dynamics rollouts let the planner anticipate the future effect of each candidate action before committing to it, whereas a VLM-based selector relies on implicit reasoning alone without such a forward model. Participants will thus rate \systemname{} higher than \textbf{VLM Planner}.

\textbf{H2:} The same limitation should hold for a VLM that selects actions directly from the observation and instruction, as it likewise has no forward model to anticipate the effect of its actions. Participants will thus rate \systemname{} higher than \textbf{End-to-End VLM}.

\textbf{H3:} Holding the planner fixed, a higher-fidelity dynamics model tracks the true dynamics more closely, steering planning toward better actions and fewer mid-trajectory mispredictions. Participants will thus rate \systemname{} higher than \textbf{PC-GNN Model}.

\begin{figure*}[t]
    \centering
    \includegraphics[width=\linewidth]{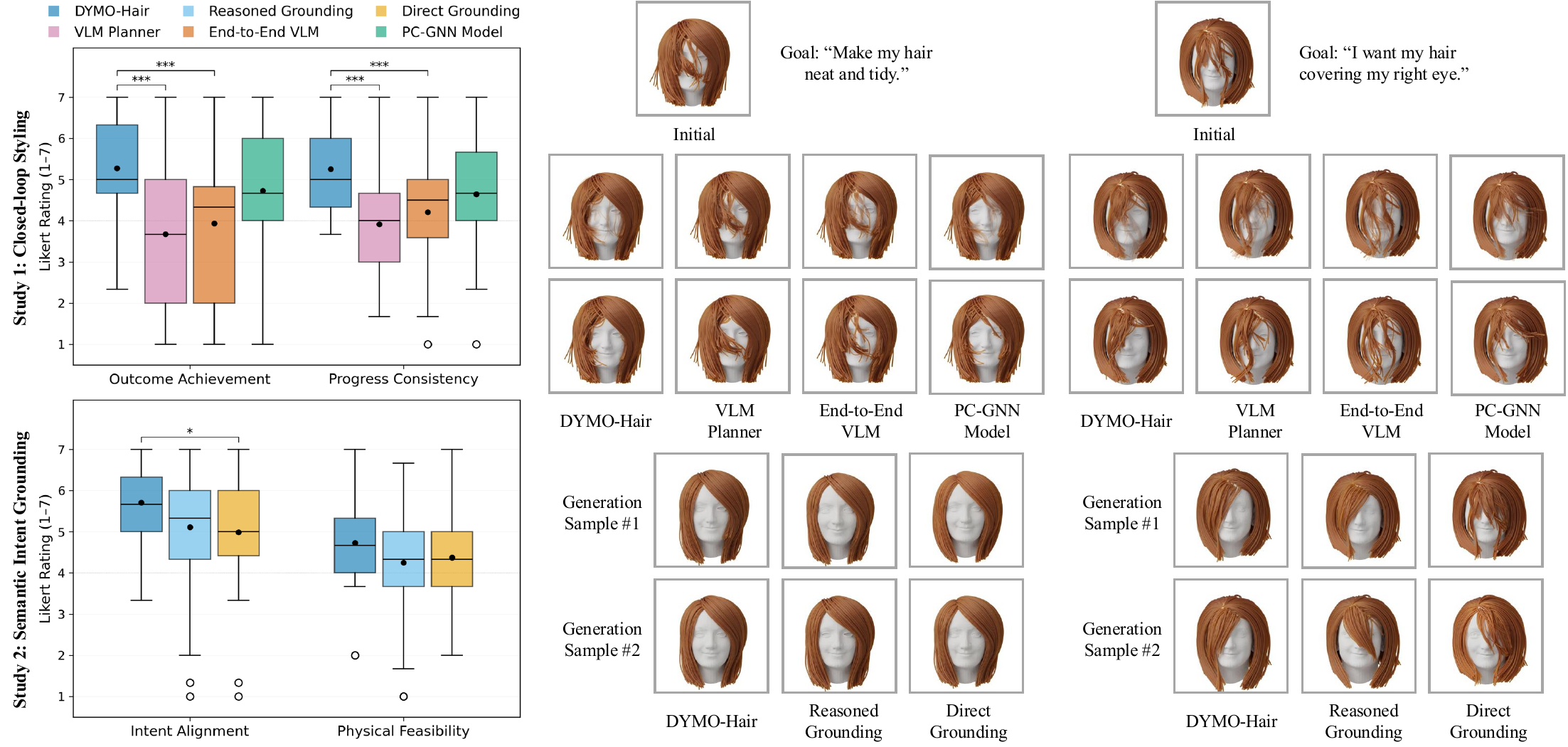}
    \caption{\textbf{Rating Results and Qualitative Examples for Language Goal-conditioned Styling.}
    \textbf{Left:} per-construct Likert scores (1--7), each the average of the construct's three items, for every method-construct cell in each study.
    Boxes span Q1--Q3 with the median line; whiskers reach $1.5\times$ the interquartile range; open circles are outliers; black dots are LME estimated marginal means.
    Brackets mark planned pairwise comparisons that are significant under the Holm-corrected LME ($^{*}p<.05$, $^{**}p<.01$, $^{***}p<.001$); all planned comparisons are reported in~\cref{tab:user-study-quant-main}.
    \textbf{Right:} two qualitative cases, each with its own hairstyle and language goal, shown in the left and right groups.
    Within each case, the initial hairstyle and goal are given at the top; the upper block shows each Study 1 method's styling trajectory, with a snapshot during planning above the final styled state, and the lower block shows two goal image samples generated by each Study 2 variant.}
    \vspace{-5pt}
    \label{fig:user-study-rating-boxplot}
\end{figure*}

\begin{table}[t]
\centering
\small
\caption{\textbf{Planned Hypothesis Pairwise Comparisons for Language Goal-conditioned Styling.} Each study is evaluated on two constructs, reported in the left and right column groups.
Method abbreviations: DYMO for \systemname{}, Reasoned for \textbf{Reasoned Grounding}, and Direct for \textbf{Direct Grounding}.
$\Delta$ is the LME-estimated difference in mean rating between the two methods in Likert points (positive: the first method rated higher), and $p$ is Holm-corrected within each study (bold: $p<.05$).}
\label{tab:user-study-quant-main}
\resizebox{\linewidth}{!}{
\begin{tabular}{llcccc}
\shline
\addlinespace[1pt]
H & Contrast
 & \multicolumn{1}{c}{$\Delta$} & \multicolumn{1}{c}{$p$}
 & \multicolumn{1}{c}{$\Delta$} & \multicolumn{1}{c}{$p$} \\
 \addlinespace[1pt]
\hline
\addlinespace[1pt]
\multicolumn{2}{l}{\textit{Study 1: Closed-loop Styling}}
 & \multicolumn{2}{c}{\textit{Outcome Achievement}}
 & \multicolumn{2}{c}{\textit{Progress Consistency}} \\
\addlinespace[2pt]
H1 & DYMO vs.\ VLM Planner    & $+1.60$ & \textbf{$<$.001}  & $+1.34$ & \textbf{$<$.001}  \\
H2 & DYMO vs.\ End-to-End VLM & $+1.34$ & \textbf{$<$.001}  & $+1.05$ & \textbf{$<$.001}  \\
H3 & DYMO vs.\ PC-GNN Model   & $+0.54$ & .078   & $+0.61$ & .056   \\
 \addlinespace[1pt]
\midrule
  \addlinespace[1pt]
\multicolumn{2}{l}{\textit{Study 2: Semantic Intent Grounding}}
 & \multicolumn{2}{c}{\textit{Intent Alignment}}
 & \multicolumn{2}{c}{\textit{Physical Feasibility}} \\
\addlinespace[2pt]
H4 & Reasoned vs.\ Direct        & $+0.13$ & 1.000  & $-0.13$ & 1.000  \\
H5 & DYMO vs.\ Reasoned & $+0.60$ & .070   & $+0.48$ & .248  \\
H6 & DYMO vs.\ Direct   & $+0.72$ & \textbf{.048}  & $+0.36$ & .642  \\
 \addlinespace[1pt]
\shline
\end{tabular}
}
\end{table}

\paragraph{Results.}
As shown in~\cref{fig:user-study-rating-boxplot} and \cref{tab:user-study-quant-main}, 
comparing model-based predictive planning against VLM-based action selection, 
\systemname{} significantly outperforms both \textbf{VLM Planner} and \textbf{End-to-End VLM} on both constructs (all $p<.001$). \textbf{H1} and \textbf{H2} are \textbf{supported}.
We found that participants most often rated \systemname{} in the 5--7 range, whereas they frequently rated both VLM variants around or below the neutral midpoint of 4, judging those trajectories as not achieving or progressing consistently toward the intended hairstyle. Notably, \textbf{VLM Planner}, which receives the grounded goal image, is rated no higher than \textbf{End-to-End VLM}, which acts directly from the instruction, suggesting that access to an explicit target is not what limits VLM-based action selection.

Comparing dynamics-model fidelity with the planner held fixed, \systemname{} improves over \textbf{PC-GNN Model} on both constructs, but neither gain reaches significance ($p=.078$ for \textit{outcome achievement}, $.056$ for \textit{progress consistency}). \textbf{H3} is not supported, though both gains trend toward significance.
We read this as directional support that dynamics-model fidelity contributes to styling performance, with an effect smaller than the gap between model-based and VLM-based planning.
Our user study results match our standalone dynamics evaluation in~\cref{sec:exp-dynamics}. We found that \textbf{PC-GNN} captures hair dynamics only partially but still gives the planner some lookahead, while \systemname{}'s model predicts local deformation more accurately, which is conducive to more effective hair styling.

\subsubsection{Study 2: Semantic Intent Grounding.}
\label{sec:user-study-grounding}
This study further assesses how faithfully the grounded goal image reflects the user's intent, and whether it is physically reachable. We compare the full grounding module against two ablations that cumulatively remove its VLM-based mediation stages, testing whether each stage improves goal-image quality. We find that pre-generation reasoning alone has no measurable effect, and that the two stages together significantly improve intent alignment, with post-generation selection converting reasoning's unstable gains into a goal image that is more consistent across samples.

\paragraph{Constructs.}
We measure two constructs, each rated on three 7-point Likert items, including one reverse-worded item (full item wording in the appendix):
\begin{itemize}[leftmargin=*,itemsep=0em]
\item \textit{Intent alignment}: how well the generated goal image reflects the hairstyle the user expressed in language.
\item \textit{Physical feasibility}: how plausibly the generated goal image can be reached from the initial hairstyle using the robot's combing-based styling capabilities.
\end{itemize}

\paragraph{Baselines.}
We compare \systemname{}'s full intent grounding module against two ablated variants:
\begin{itemize}[leftmargin=*,itemsep=0em]
\item \textbf{Direct Grounding}: removes the VLM-based mediation entirely, conditioning the image generator directly on the current observation and instruction, with no pre-generation reasoning or post-generation selection.

\item \textbf{Reasoned Grounding}: removes only post-generation selection, retaining pre-generation reasoning to interpret the instruction into a concrete styling plan, then generating a single goal image conditioned on that plan.
\end{itemize}

\paragraph{Hypotheses.}
We examine the two VLM-based mediation stages, added cumulatively: \textbf{H4} and \textbf{H5} test each stage as it is introduced, and \textbf{H6} tests both together against the unmediated baseline. Each hypothesis predicts higher goal-image quality, measured by the two constructs above.

\textbf{H4:} Pre-generation reasoning turns an under-specified instruction into a concrete styling plan, giving the image generator a target that is sharper in intent and aware of the robot's capabilities. Participants will thus rate \textbf{Reasoned Grounding} higher than \textbf{Direct Grounding}.

\textbf{H5:} Post-generation selection keeps, from multiple generated images, the one most consistent with user intent and robot capability. Participants will thus rate \systemname{} higher than \textbf{Reasoned Grounding}.

\textbf{H6:} The two stages are complementary: reasoning broadens the space of candidate goal images and selection retains the best, together yielding a better-grounded target than the unmediated baseline. Participants will thus rate \systemname{} higher than \textbf{Direct Grounding}.

\paragraph{Results.}
As shown in~\cref{fig:user-study-rating-boxplot} and \cref{tab:user-study-quant-main}, we find no support for \textbf{H4}: pre-generation reasoning alone does not help, as \textbf{Reasoned Grounding} is not significantly different from \textbf{Direct Grounding} on either construct (all $p=1.000$). \textbf{H5} is also not supported: with post-generation selection added, \textbf{\systemname{}} is rated higher than \textbf{Reasoned Grounding} on both constructs, but neither reaches significance, though the \textit{intent alignment} gain trends toward significance ($p=.070$). \textbf{H6} is \textbf{partially supported}: combining the two stages significantly improves \textit{intent alignment} over \textbf{Direct Grounding} ($p=.048$), while the \textit{physical feasibility} gain remains non-significant ($p=.642$).

The rating distributions suggest why. Reasoning alone does not raise the average goal quality but makes it more variable: relative to \textbf{Direct Grounding}, the spread of ratings widens (standard deviation, SD: $1.36{\to}1.48$ for \textit{intent alignment} and $1.24{\to}1.43$ for \textit{physical feasibility}). It yields better goals on some samples but not others, so no average gain emerges. Selection converts this into a reliable improvement by keeping the best of several candidates: it raises the mean \textit{intent alignment} rating ($5.07{\to}5.69$) while sharply reducing the spread ($1.48{\to}0.92$). The two stages are thus complementary: reasoning raises the quality that can be reached, and selection secures it by discarding the misses, yielding a goal image that is both better and more consistent.

\section{Experiments on Closed-loop Hair Styling in the Real World}
\label{sec:exp-real-world-closed-loop}
We further evaluate how well \systemname{} transfers from simulation to the real world, using the simulation-trained dynamics model directly without any real-world fine-tuning.

\vspace{-5pt}

\subsection{Generalization across Hair Types}
\begin{figure*}[t]
\centering
    \includegraphics[width=\linewidth]{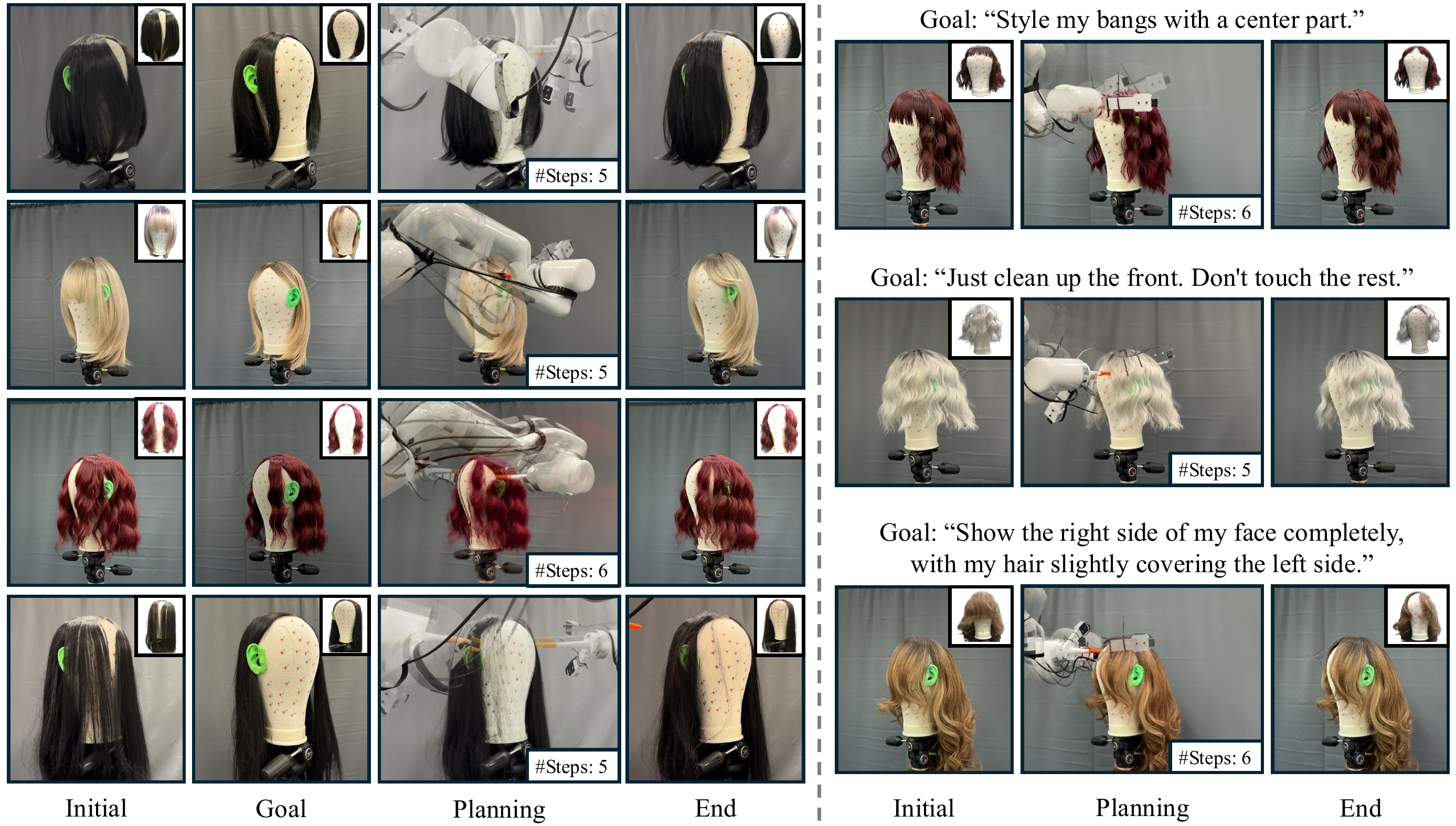}
    \caption{\textbf{Real-world Hair Styling across Different Hair Types.}
    Zero-shot styling results of \systemname{} on 7 physical wigs spanning varied colors, lengths, and textures, under both visual and language goals, with no real-world fine-tuning.
    \textbf{Left: visual goal-conditioned styling.} Columns show, from left to right, the initial state, the goal state, a snapshot of the intermediate planning steps, and the styled end state.
    \textbf{Right: language goal-conditioned styling.} The language goal is given above each example, followed by the initial state, a planning snapshot, and the styled end state.
    Insets in the top-right corner show the front view of each hair state, and each planning snapshot is annotated with the total number of planning steps.}
    \label{fig:exp-real-qual-gallery}
\end{figure*}

We first test \systemname{} on 7 physical wigs spanning four colors (black, white, red, and blonde), three lengths (short, medium, and long), and two textures (straight and wavy), under both visual and language goals.
As shown in~\cref{fig:exp-real-qual-gallery}, \systemname{} styles each wig effectively toward its specified goal in a zero-shot manner, generalizing across different hair colors, lengths, and textures.

\vspace{-5pt}

\subsection{Visual Goal-conditioned Styling versus Rule-based System}
\begin{figure*}[t]
    \centering
    \includegraphics[width=\linewidth]{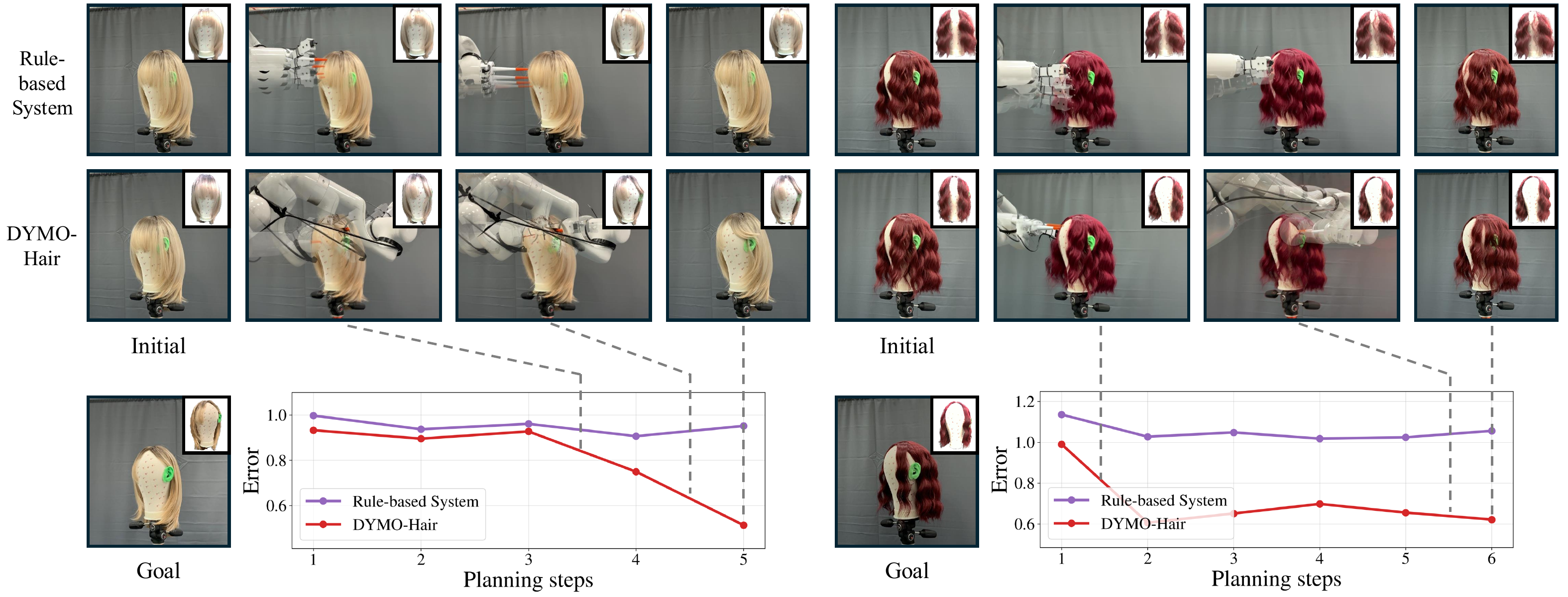}
    \caption{\textbf{Real-world Visual Goal-conditioned Styling: \systemname{} vs.\ Rule-based System.}
    For each case, the two methods are shown row by row from a nearly identical initial hair state.
    Each row shows selected keyframes of the styling trajectory, with insets in the top-right corner giving the front view of the hair after the action in each keyframe.
    The shared visual goal and the per-step error curve for both methods are shown at the bottom of each case; dashed lines connect each keyframe to its corresponding point on the error curve.
    Error is the relative strand-level distance to the goal, normalized by the initial distance; step 0 is omitted, as this relative error is 1 by definition at the start.}
    \label{fig:exp-real-quant-closed-loop-visual}
\end{figure*}

We compare \systemname{} against \textbf{Rule-based System} (\cref{sec:exp-sim-closed-loop-visual}), the only existing visual goal-conditioned hair styling system. As shown in~\cref{fig:exp-real-quant-closed-loop-visual}, \systemname{} reduces the geometric error overall and effectively styles the hair toward the specified goal, whereas \textbf{Rule-based System} fails in both cases. 

Leveraging its dynamics model trained in simulation, \systemname{} plans actions predicted to move the hair toward the goal at each step. However, real-world physical challenges, such as unconstrained hair fall under gravity and complex contact around the ear, mean a planned deformation does not always play out as intended (left blonde hair case, steps 1–3; right red hair case, steps 3–4). The dynamics model and closed-loop planning together mitigate this gap: after each action, \systemname{} re-estimates the actual hair state and replans from it, searching with the dynamics model for a corrective action grounded in what actually happened; repeated correction then drives the hair toward the goal even when individual steps fall short (left case, steps 4–5; right case, steps 5–6).

Although \textbf{Rule-based System} also plans in closed loop, its replanning often fails to converge because its handcrafted rules are brittle, for two main reasons. First, the rules compute geometric differences through strand-level correspondence, which requires accurate 2D orientation maps and strand tracking. 
These break down as errors accumulate on longer hair and, as we observed during experiments, under varying hair color and appearance, where specular reflections or poor contrast can corrupt the 2D orientation estimates. 
Second, the system plans from a single front-view observation, which cannot capture changes that extend to the side. In the left blonde hair case in~\cref{fig:exp-real-quant-closed-loop-visual}, for instance, success requires pushing hair behind the left ear, a change only partly visible from the front.
In contrast, \systemname{} estimates the 3D hair state from multiple views and evaluates geometric differences without strand-level correspondence, making it more robust to variation in hair color and appearance, better able to generalize across hair lengths, and able to capture goals that span multiple views. Its dynamics model, in turn, supplies reliable future state predictions for planning, which lets closed-loop replanning converge rather than merely repeat.

\begin{figure*}[t]
\centering
    \includegraphics[width=\linewidth]{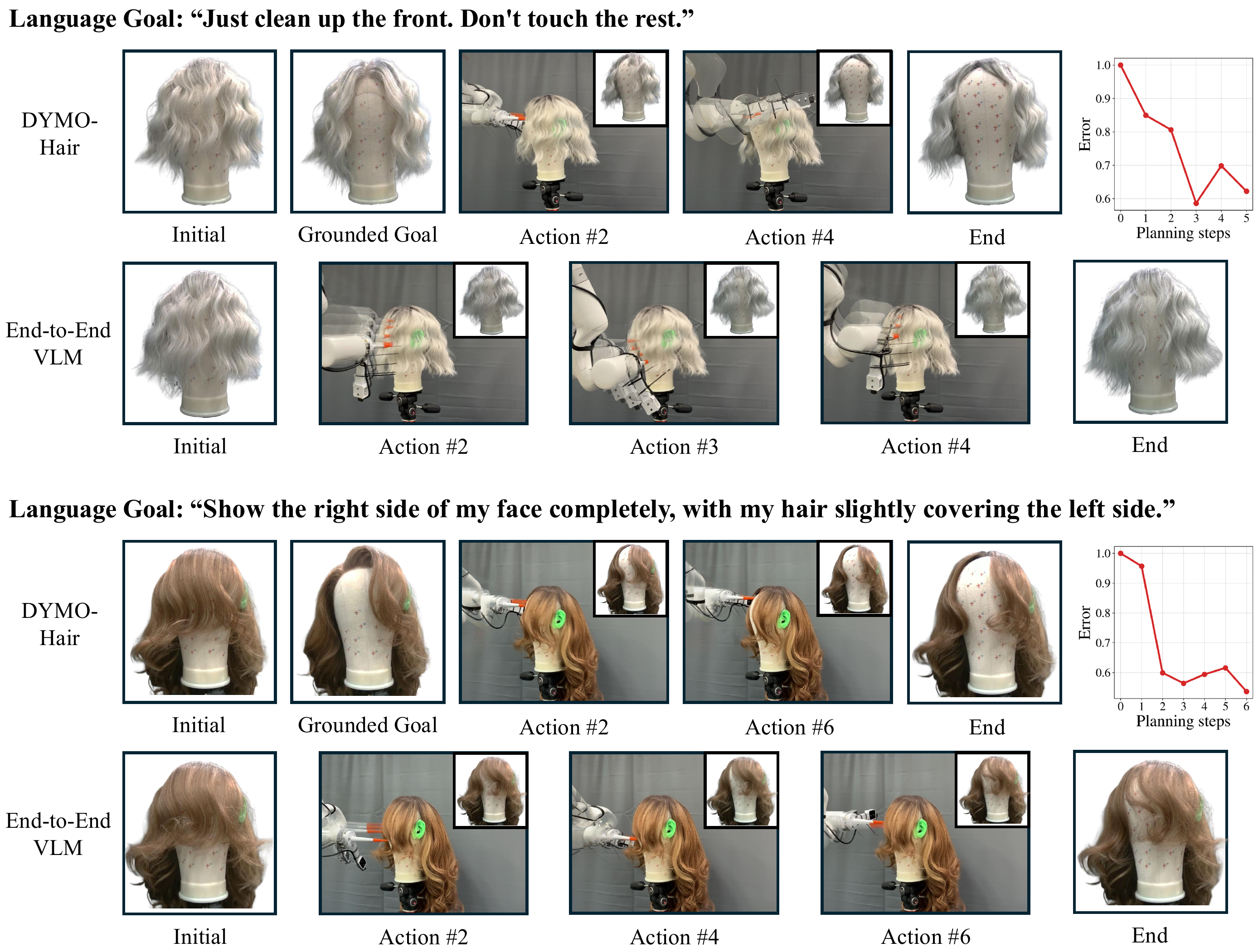}
    \caption{\textbf{Real-world Language Goal-conditioned Styling: \systemname{} vs.\ End-to-End VLM.}
    For each case, the language goal is given at the top and each method is shown in one row.
    Within a row, we show the initial state, selected keyframes from the styling trajectory, and the final result; insets in the top-right corner give the front view of the hair after the action in each keyframe.
     \systemname{} first grounds the instruction into a \textbf{Grounded Goal} image, which enables closed-loop planning against an explicit target, and the rightmost plot reports the resulting relative error over planning steps. The End-to-End VLM has no such target and instead predicts actions directly from the language goal.}
    \label{fig:exp-real-qual-language}
 \end{figure*}

\subsection{Language Goal-conditioned Styling versus VLM-based Baseline}
For language goal-conditioned styling, we compare \systemname{} against \textbf{End-to-End VLM} (\cref{sec:user-study-styling}), in which a pretrained VLM (Gemini 3.1 Pro~\citep{gemini31pro2026}) takes the current observation and the language instruction and selects an action directly from a set of sampled candidates via PIVOT~\citep{nasiriany2024pivot}-style visual prompting.
In both cases shown in~\cref{fig:exp-real-qual-language}, \systemname{} first grounds the language intent into a goal image and then plans and styles the hair toward it, with the error decreasing overall as the hair approaches the target. \textbf{End-to-End VLM} can generally interpret the instruction from its strong pretrained prior and propose plausible actions at some steps, but this prior is hard to ground accurately in physical action: reasoning over 2D action overlays in the visual prompting, it cannot reliably anticipate how each motion will reshape the hair, so although it makes progress at some steps, it also stalls or drifts away from the goal.

Both behaviors mirror our findings from the simulation user study (\cref{sec:exp-sim-closed-loop-language}) and reflect the same underlying mechanism. \systemname{} selects each action through predictive rollouts of its dynamics model, explicitly anticipating the physical consequence of a comb motion before committing to it, whereas \textbf{End-to-End VLM} relies on implicit reasoning alone, without such a forward model. As that study established, this advantage comes not from simply supplying a grounded target image, since the VLM struggled with or without one, but from the lookahead that predictive planning affords.

\section{Conclusion}
\label{sec:conclusion}

This paper presents a study on model-based robot hair manipulation. We introduce a 3D hair-combing dynamics model, built on a novel volumetric learning method for generalizable dynamics modeling. To support the data requirements of large-scale pre-training, we further develop a GPU-parallelized PBD-based simulator tailored for strand-level, contact-rich hair-combing simulation. Together, these components yield \systemname{}, a model-based robot hair care system for multimodal goal-conditioned hair styling from either visual references or natural language instructions. 
Evaluated in both simulation and the real world, \systemname{} generalizes to unseen hairstyles and transfers zero-shot from simulation to the real world. These results lay a foundation for more generalizable, flexible, and accessible robot hair care, and more broadly, for assistive robots that can respond to diverse user goals and support people throughout their daily lives.

Several limitations remain for future work. 
First, we focus on model-based planning with limited treatment of human-robot interaction, such as enforcing action-space constraints that keep the tool away from safety-critical regions like the eyes to enhance real-world usability. 
Second, the system assumes privileged mannequin head information and perfect calibration; incorporating online head estimation would improve practicality. 
Third, our evaluation covers straight and wavy hair but not tightly curly or coily hair, whose contact and clumping behavior differs substantially from the straight and wavy hair the simulator models; without such synthetic dynamics data, the dynamics model is not expected to generalize to these types, which we leave as an important future direction for broader accessibility.
Finally, we use a simple 3D-printed combing tool in the real-world system to mirror the simulation setup; replacing it with more advanced designs such as soft robot fingers~\citep{yoo_2025_moehair, yoo2025kinesoft} could further enhance usability.

\begin{acks}
We would like to thank Feiyu Zhu and John Z. Zhang for their valuable feedback and discussions.
\end{acks}

\begin{funding}
This work was supported by NSF IIS-2112633 and NSF Graduate Research Fellowship under Grant No. DGE2140739.
\end{funding}

\bibliographystyle{SageH}
\bibliography{reference}

\appendix
\section{User Study Details}

\subsection{Procedure}

Participants completed Study 1 before Study 2.

For Study 1, each participant viewed and rated 4 demonstrations, sampled uniformly from the 96 demonstrations spanning \systemname{} and the three planning baselines ($4 \text{ methods} \times 8 \text{ tasks} \times 3 \text{ seeds} $). 
Each demonstration presented the initial observation, the language instruction, and a 5-image sequence of the styling trajectory: 4 intermediate keyframes uniformly spaced across execution, followed by the final styled hairstyle.
Participants rated it on \textit{outcome achievement} and \textit{progress consistency}, each measured by three 7-point Likert items.

For Study 2, each participant viewed and rated 2 demonstrations, sampled uniformly from the 72 demonstrations spanning \systemname{} and its two ablated variants ($3 \times 8 \times 3 $ ). Each demonstration presented the initial observation, the language instruction, and the goal image produced by the grounding module. Participants rated it on \textit{intent alignment} and \textit{physical feasibility}, each also measured by three 7-point Likert items.

Sampling was uniform rather than explicitly balanced, but the resulting allocation was close to even: each method received between 37 and 41 in Study 1 and 34 and 45 ratings in Study 2.

Since demonstrations were sampled independently rather than matched across methods, participants did not systematically rate the same task under different methods. Method comparisons are therefore made at the population level. In the primary LME analysis, the participant random intercept accounts for differences in individual rating tendency.

\subsection{Questionnaire Design}
\label{sec:user-study-questionnaire}
Each construct was measured by three 7-point Likert items (\textit{strongly disagree} to \textit{strongly agree}), one of which was reverse-worded and reverse-scored before averaging. 
The full item wording is given below; bracketed labels indicate each item's target construct and index, and were not shown to participants.

\paragraph{Study 1.}
The following items were shown under the heading \textit{``Comparing the language instruction with the final achieved hairstyle (the `Final' image), I found that\ldots''}:
\begin{itemize}[leftmargin=*,itemsep=0em]
\item The final hairstyle reflects the meaning of the instruction. [\textit{outcome achievement}, item 1]

\item The intended hairstyle in the instruction is achieved in the final result. [\textit{outcome achievement}, item 2]

\item The final hairstyle does not satisfy the instruction. [\textit{outcome achievement}, item 3, reverse-scored]
\end{itemize}

The following items were shown under the heading \textit{``Comparing the language instruction with all the frames (the `Keyframe 1--4' and `Final' images), I found that\ldots''}:
\begin{itemize}[leftmargin=*,itemsep=0em]
\item Across the frames, the hairstyle steadily moves toward the instructed goal. [\textit{progress consistency}, item 1]

\item Each step of the styling process moves the hair closer to the intended goal of the instruction. [\textit{progress consistency}, item 2]

\item Some steps undo or reverse earlier progress toward the goal. [\textit{progress consistency}, item 3, reverse-scored]
\end{itemize}

\paragraph{Study 2.}
The following items were shown under the heading \textit{``Comparing the generated goal hairstyle with the language instruction, I found that\ldots''}:
\begin{itemize}[leftmargin=*,itemsep=0em]
\item The goal hairstyle matches what the instruction asked for. [\textit{intent alignment}, item 1]

\item The goal hairstyle captures the hairstyle implied by the instruction. [\textit{intent alignment}, item 2]

\item The goal hairstyle suggests a different intent than the instruction. [\textit{intent alignment}, item 3, reverse-scored]

\item The goal hairstyle could be achieved from the starting hair state just by combing. [\textit{physical feasibility}, item 1]

\item The goal hairstyle does not require hairstyling beyond combing. [\textit{physical feasibility}, item 2]

\item Achieving the goal hairstyle from the starting hair state would require more than combing. [\textit{physical feasibility}, item 3, reverse-scored]
\end{itemize}

\subsection{Internal Consistency Check}
\label{sec:user-study-alpha-reliability}
Since each construct is measured by averaging three Likert items, it is worth checking how consistently those items behaved. Cronbach's $\alpha$ summarizes, on a $0$--$1$ scale, how strongly the items of a construct agree with one another: high values indicate that participants rated a demonstration similarly across all three items, while low values indicate that at least one item behaved differently from the rest.

\cref{tab:user-study-cronhach-alpha} reports Cronbach's $\alpha$ for each construct together with the pairwise correlations among its three items. Internal consistency is high for \textit{outcome achievement} ($\alpha=0.84$) and more modest for the other three constructs ($\alpha=0.61$, $0.69$, and $0.64$). In every case the lower coefficients trace to the reverse-worded item: the two positively worded items correlate strongly with each other ($r_{1,2}=0.63$--$0.89$), whereas each correlates considerably less with the reverse-worded one ($r_{1,3}$ and $r_{2,3}$).

The effect is most pronounced for \textit{progress consistency}, where the reverse-worded item correlates only weakly with the other two, suggesting it captured the construct less well than intended. We suspect this reflects the difficulty of writing reverse-worded items for constructs of this kind: styling quality is broad and hard to pin down precisely, so a reverse-worded statement that reads as reasonable on its own may not be a clean negation of the positive items. For \textit{progress consistency}, for example, a trajectory can move toward the goal well overall while still containing a few steps that undo earlier progress, due to sampling stochasticity in the planner or occasional dynamics-model mispredictions, so the two framings are not strict opposites.

However, a weakly correlated item can only add noise to the averaged score: it inflates residual variance and widens standard errors, working against the detection of differences rather than creating them. The pairwise comparisons on these three constructs are therefore conservative, and the large effects on \textit{progress consistency} against both VLM baselines (\textbf{H1}, \textbf{H2}) were detected despite this attenuation. Restricting the construct to its two positively worded items leaves both the conclusions and the significance of \textbf{H1} and \textbf{H2} unchanged.

\begin{table}[t]
\centering
\small
\caption{\textbf{Internal Consistency of the Rating Constructs.} Cronbach's $\alpha$ is computed over all three items of each construct. Items 1 and 2 are positively worded and item 3 is reverse-worded and reverse-scored (listed in~\cref{sec:user-study-questionnaire}). $r$ values are Pearson correlations between item pairs, and $n$ is the number of rated demonstrations.}
\label{tab:user-study-cronhach-alpha}
\begin{tabular*}{0.95\linewidth}{@{\extracolsep{\fill}}lccccc@{}}
\toprule
Construct & $\alpha$ & $r_{1,2}$ & $r_{1,3}$ & $r_{2,3}$ & $n$ \\
\midrule
\multicolumn{6}{l}{\textit{Study 1: Closed-loop Styling}} \\
\addlinespace[1pt]
\textit{Outcome achievement}  & 0.84 & 0.88 & 0.54 & 0.54 & 232 \\
\textit{Progress consistency} & 0.61 & 0.89 & 0.13 & 0.08 & 232 \\
\midrule
\multicolumn{6}{l}{\textit{Study 2: Semantic Intent Grounding}} \\
\addlinespace[1pt]
\textit{Intent alignment}     & 0.69 & 0.81 & 0.29 & 0.28 & 116 \\
\textit{Physical feasibility} & 0.64 & 0.63 & 0.29 & 0.23 & 116 \\
\bottomrule
\end{tabular*}
\end{table}

\subsection{Analysis Protocol}
\label{sec:appendix-user-study-analysis-protocol}

\paragraph{Linear Mixed-effects (LME) Models.}
For each construct, we fit a separate model to the averaged Likert score, using restricted maximum likelihood (REML) estimation via the \texttt{pymer4} Python package. The full model is
\begin{equation*}
\texttt{score} \sim \texttt{method} + (1 \mid P) + (1 \mid T) + (1 \mid D),
\end{equation*}
where \texttt{method} is a categorical fixed effect with \systemname{} as the reference level, and $P$, $T$, and $D$ are random intercepts capturing three sources of non-independence in the study design:
\begin{itemize}[leftmargin=*,itemsep=0em]
\item participant ($P$): one participant's ratings are correlated through their individual rating tendency;
\item task ($T$): ratings of demonstrations from the same task share that task's difficulty;
\item demonstration ($D$): the same generated demonstration is rated by several participants, and each task is repeated under several seeds, so demonstration is nested within but not identical to task.
\end{itemize}

A random-effect variance can be estimated at or near zero, producing a singular fit, either because the grouping genuinely accounts for little variance or because too few observations are available to resolve it. We therefore fit a fallback ladder of progressively simpler models and take the first that converges without singularity:
\begin{enumerate}[leftmargin=*,itemsep=0em]
\item $\texttt{score} \sim \texttt{method} + (1 \mid P) + (1 \mid T) + (1 \mid D)$ (the full model)
\item $\texttt{score} \sim \texttt{method} + (1 \mid P) + (1 \mid D)$
\item $\texttt{score} \sim \texttt{method} + (1 \mid P) + (1 \mid T)$
\item $\texttt{score} \sim \texttt{method} + (1 \mid P)$
\end{enumerate}
The participant intercept is retained throughout, since every rating belongs to a participant and each participant contributes several ratings. If the last model is also singular, it is used regardless.

Pairwise comparisons between methods are then estimated with \texttt{emmeans} on the fitted model, giving for each comparison an effect size $\Delta$ in Likert points and its statistical significance. We apply Holm correction across the six planned pairwise comparisons within each study (three method pairs $\times$ two constructs) and report only these planned comparisons.

\paragraph{Non-parametric Check.}
Since the LME assumes approximately normal residuals, which Likert scores may violate, we additionally evaluate each planned pairwise comparison with a Wilcoxon signed-rank test as a distribution-free check. For a comparison between method A and B, every participant who rated both methods contributes one paired difference: the mean of their scores for A minus the mean of their scores for B, averaged over whichever demonstrations they happened to view. 
The test is two-sided, and we report the matched-pairs rank-biserial correlation $r$ as its effect size, which ranges from $-1$ to $+1$ and is positive when most participants rated method A higher.
Holm correction is applied across the same six planned pairwise comparisons per study as for the LME.

\paragraph{Interpreting the Two Tests.}
The two tests trade off against each other: the LME models the nested rating structure but assumes normally distributed residuals, while the non-parametric Wilcoxon test makes no distributional assumption but ignores that structure. Agreement therefore indicates a conclusion that does not depend on either limitation.

The Wilcoxon test is, however, substantially less powerful than the LME. Since demonstrations for rating were sampled independently rather than matched across methods, only participants who happened to rate both methods of a pairwise comparison contribute a paired difference, far fewer than the ratings available to the LME. We therefore treat the check as evidence about the direction and robustness of an effect rather than as an independent significance test, and a pairwise comparison that is significant under the LME but not under the Wilcoxon test is expected where the effect is moderate.

\begin{table}[t]
\centering
\small
\caption{\textbf{Likert Rating Distributions.} Mean and standard deviation (SD) of the averaged three-item Likert score (1--7) for each method-construct cell, pooled across tasks, random seeds, and participants.}
\label{tab:user-study-construct-stats}
\begin{tabular*}{\linewidth}{@{\extracolsep{\fill}}lcccc@{}}
\toprule
\multicolumn{5}{l}{\textit{Study 1: Closed-loop Styling}} \\
\addlinespace[2pt]
& \multicolumn{2}{c}{\textit{Outcome Achi.}} & \multicolumn{2}{c}{\textit{Progress Cons.}} \\
\cmidrule(lr){2-3}\cmidrule(lr){4-5}
Method & \multicolumn{1}{c}{Mean} & \multicolumn{1}{c}{SD} & \multicolumn{1}{c}{Mean} & \multicolumn{1}{c}{SD} \\
\midrule
VLM Planner    & 3.50 & 1.84 & 3.86 & 1.27 \\
End-to-End VLM & 3.98 & 1.66 & 4.22 & 1.47 \\
PC-GNN Model   & 4.73 & 1.55 & 4.66 & 1.36 \\
\systemname{}  & 5.30 & 1.16 & 5.23 & 0.92 \\
\bottomrule

\vspace{0.8em}

\end{tabular*}
\begin{tabular*}{\linewidth}{@{\extracolsep{\fill}}lcccc@{}}
\toprule
\multicolumn{5}{l}{\textit{Study 2: Semantic Intent Grounding}} \\
\addlinespace[2pt]
& \multicolumn{2}{c}{\textit{Intent Alig.}} & \multicolumn{2}{c}{\textit{Physical Feas.}} \\
\cmidrule(lr){2-3}\cmidrule(lr){4-5}
Method & \multicolumn{1}{c}{Mean} & \multicolumn{1}{c}{SD} & \multicolumn{1}{c}{Mean} & \multicolumn{1}{c}{SD} \\
\midrule
Direct Grounding   & 5.03 & 1.36 & 4.45 & 1.24 \\
Reasoned Grounding & 5.07 & 1.48 & 4.22 & 1.43 \\
\systemname{}      & 5.69 & 0.92 & 4.69 & 1.14 \\
\bottomrule
\end{tabular*}

\end{table}

\subsection{Results}
\label{sec:appendix-user-study-results}

\cref{tab:user-study-construct-stats} reports the per-method Likert rating distribution, \ie, the mean and standard deviation of each construct. \systemname{} has the lowest spread on every construct, indicating that its ratings are more consistent across samples.

\cref{tab:appendix-lme-spec} reports the random-effect structure selected for each construct by the fallback ladder. The selected structures differ across constructs, as variance components that cannot be resolved from the available ratings are dropped by the procedure described in~\cref{sec:appendix-user-study-analysis-protocol}.

\cref{tab:appendix-user-study-full-contrasts} reports the full analysis results for all planned pairwise comparisons. The two tests agree in direction on eleven of the twelve comparisons. The exception is \textbf{H4} on \textit{physical feasibility}, where the LME $\Delta$ is negative while the Wilcoxon rank-biserial $r$ is positive; both tests are far from significance there ($p_{\text{Holm}}=1.000$ under each), so the disagreement concerns a pairwise comparison for which neither analysis detects an effect.

The paired sample sizes for the Wilcoxon check vary considerably across comparisons, from 9 to 18 participants, against the 232 and 116 ratings per construct available to the LME in Study 1 and Study 2 respectively. Where the paired sample is larger, the check corroborates the primary analysis: the four Study 1 pairwise comparisons against the VLM baselines draw on 16 to 18 participants, three of which remain significant under the non-parametric check and the fourth borderline ($p_{\text{Holm}}=.060$). Where the paired sample is smallest, the check has little power to detect an effect. This is the case for \textbf{H6} on \textit{intent alignment}: the LME finds a significant difference over 116 ratings, while the paired test draws on 9 participants and returns $p_{\text{raw}}=1.000$ before any correction. The non-parametric outcome there reflects the small paired sample rather than evidence against the effect.

\begin{table}[t]
\centering
\small
\caption{\textbf{LME Model-fitting Specification.} For each construct we report the random-effect structure selected by the fallback ladder ($P$: participant, $T$: task, $D$: demonstration) and the number of ratings $n$ it was fit to. The fixed effect \texttt{method} is omitted since it is included in every model. All selected models converged without singularity.}
\label{tab:appendix-lme-spec}
\begin{tabular*}{\linewidth}{@{\extracolsep{\fill}}llr@{}}
\toprule
Construct & Random effects & \multicolumn{1}{c}{$n$} \\
\midrule
\multicolumn{3}{l}{\textit{Study 1: Closed-loop Styling}} \\
\addlinespace[1pt]
\textit{Outcome achievement}  & $(1 \mid P) + (1 \mid D)$ & 232 \\
\textit{Progress consistency} & $(1 \mid P) + (1 \mid D)$ & 232 \\
\midrule
\multicolumn{3}{l}{\textit{Study 2: Semantic Intent Grounding}} \\
\addlinespace[1pt]
\textit{Intent alignment}     & $(1 \mid P) + (1 \mid T) + (1 \mid D)$ & 116 \\
\textit{Physical feasibility} & $(1 \mid P)$                          & 116 \\
\bottomrule
\end{tabular*}
\end{table}

\begin{table*}[t]
\centering
\small
\caption{\textbf{Full Planned Pairwise Comparison Results for Language Goal-conditioned Styling.} $\Delta$ is the LME effect-size estimate in Likert points (positive: the first method rated higher). $r$ is the matched-pairs rank-biserial correlation from the participant-paired Wilcoxon signed-rank test, computed over the $n$ participants who rated both methods of the comparison. For each test we report the uncorrected $p$-value ($p_{\text{raw}}$) and the Holm-corrected value ($p_{\text{Holm}}$), corrected across the six planned comparisons within each study. Significant values ($p_{\text{Holm}}<.05$) are in bold.}
\label{tab:appendix-user-study-full-contrasts}
\begin{tabular*}{\linewidth}{@{\extracolsep{\fill}}llrrrrrrc@{}}
\toprule
& & \multicolumn{3}{c}{LME} & \multicolumn{4}{c}{Non-parametric Wilcoxon} \\
\cmidrule(lr){3-5}\cmidrule(lr){6-9}
H & Pairwise Comparison & \multicolumn{1}{c}{$\Delta$} & \multicolumn{1}{c}{$p_{\text{raw}}$} & \multicolumn{1}{c}{$p_{\text{Holm}}$} & \multicolumn{1}{c}{$r$} & \multicolumn{1}{c}{$p_{\text{raw}}$} & \multicolumn{1}{c}{$p_{\text{Holm}}$} & $n$ \\
\midrule
\multicolumn{9}{l}{\textit{Study 1: Closed-loop Styling} --- \textit{Outcome Achievement}} \\
\addlinespace[1pt]
H1 & \systemname{} vs.\ VLM Planner    & $+1.60$ & $<$.001 & \textbf{$<$.001} & $+0.88$ & .010 & \textbf{.039} & 16 \\
H2 & \systemname{} vs.\ End-to-End VLM & $+1.34$ & $<$.001 & \textbf{$<$.001} & $+0.97$ & .003 & \textbf{.017} & 18 \\
H3 & \systemname{} vs.\ PC-GNN Model   & $+0.54$ & .078 & .078 & $+0.50$ & .088 & .099 & 16 \\
\addlinespace[3pt]
\multicolumn{9}{l}{\textit{Study 1: Closed-loop Styling} --- \textit{Progress Consistency}} \\
\addlinespace[1pt]
H1 & \systemname{} vs.\ VLM Planner    & $+1.34$ & $<$.001 & \textbf{$<$.001} & $+0.76$ & .020 & .060 & 16 \\
H2 & \systemname{} vs.\ End-to-End VLM & $+1.05$ & $<$.001 & \textbf{$<$.001} & $+0.78$ & .006 & \textbf{.030} & 18 \\
H3 & \systemname{} vs.\ PC-GNN Model   & $+0.61$ & .028 & .056 & $+0.67$ & .050 & .099 & 16 \\

\midrule

\multicolumn{9}{l}{\textit{Study 2: Semantic Intent Grounding} --- \textit{Intent Alignment}} \\
\addlinespace[1pt]
H4 & Reasoned Grounding vs.\ Direct Grounding       & $+0.13$ & .627 & 1.000 & $+0.53$ & .184 & .734  & 11 \\
H5 & \systemname{} vs.\ Reasoned Grounding & $+0.60$ & .014 & .070  & $+0.79$ & .009 & .054  & 18 \\
H6 & \systemname{} vs.\ Direct Grounding   & $+0.72$ & .008 & \textbf{.048} & $+0.05$ & 1.000 & 1.000 & 9 \\
\addlinespace[3pt]
\multicolumn{9}{l}{\textit{Study 2: Semantic Intent Grounding} --- \textit{Physical Feasibility}} \\
\addlinespace[1pt]
H4 & Reasoned Grounding vs.\ Direct Grounding       & $-0.13$ & .650 & 1.000 & $+0.25$ & .578 & 1.000 & 11 \\
H5 & \systemname{} vs.\ Reasoned Grounding & $+0.48$ & .062 & .248  & $+0.57$ & .053 & .266  & 18 \\
H6 & \systemname{} vs.\ Direct Grounding   & $+0.36$ & .214 & .642  & $+0.36$ & .379 & 1.000 & 9 \\
\bottomrule
\end{tabular*}
\end{table*}

\section{Simulation Details}
\label{sec:appendix-simulation}

Each hairstyle from USC-HairSalon~\citep{hu2015single} contains 10K strands, with each strand represented by 100 consecutive 3D points. For efficient simulation, we downsample each hairstyle to 2K strands while preserving its overall geometry. To improve simulation stability, we further preprocess each strand to ensure appropriate linear density while maintaining its curvature.

Directly importing these hairstyles into the simulator can lead to physical incompatibilities, since they were originally designed for visual realism rather than physical accuracy. In particular, they typically do not account for physical properties such as the need for inner or external forces to preserve a hairstyle in reality. As a result, hair may fail to maintain its original shape and instead sag under gravity at the start of simulation.  
To mitigate this issue, we introduce a gravity scheduling mechanism in the hair-combing simulation. Specifically, gravity is enabled only for strands that exhibit sufficient positional movement, typically caused by contact with the combing tool or inter-strand interactions, and disabled for all other strands. This design reflects the observation that, during combing, only strands subject to external contact forces should move, while others should remain stable. 
With this scheduling, strands influenced by external contact forces exhibit realistic motion under gravity in simulation, while the rest remain stable. Despite its simplicity, this mechanism effectively reduces gravity-induced sagging and enables visually-realistic, physically-plausible hair-combing simulations with the lightweight simulator, supporting the large-scale synthetic data generation for dynamics learning and closed-loop experiments for system evaluation.

\begin{table}[t]
\centering
\caption{\textbf{Key Simulation Parameters for Hair Combing}}
\label{table:sim-param}
\resizebox{0.48\textwidth}{!}{
\begin{tabular}{l|c|l|c}
\shline
\textbf{Parameter} & \textbf{Value} & \textbf{Parameter} & \textbf{Value} \\
\hline
Time step & 0.01 & Substeps & 5 \\
Gravity & [0, 0, -5.0] & Damping & 0.99 \\
Particle size & 0.01 & Linear Density & 1.0 \\
Stretch solver iterations & 20 & Bending solver iterations & 20 \\
Twist solver iterations & 20 & Stretch compliance & 0.0 \\
Stretch relaxation & 1.0 & Bending compliance & $1\times10^{-5}$ \\
Bending relaxation & 1.0 & Twist compliance & $1\times10^{-5}$ \\
Twist relaxation & 1.0 & Index gap $k$ (twist) & 10 $\sim$ 30 \\
\shline
\end{tabular}
}
\end{table}

The key parameters used in simulation are listed in~\cref{table:sim-param}. The index gap $k$ in the twist constraints is tuned individually for each hairstyle.

\section{Dynamics Learning Details}

\subsection{Method Implementation}
\label{sec:appendix-dynamics-implementation}

\textbf{The Proposed Method.}
In the state encoding path, the model takes a volumetric initial state of resolution $\mathcal{V}_0 = 64 \times 64 \times 128$ as input, concatenates the occupancy and orientation channels into a $\mathcal{V}_0 \times 4$ tensor, and hierarchically encodes it into lower resolutions using two 3D convolutional encoders with down-sampling. This produces latent embeddings $h_{\text{bottom}} \in \mathbb{R}^{\mathcal{V}_1 \times D_1}$ and $h_{\text{top}} \in \mathbb{R}^{\mathcal{V}_2 \times D_2}$, where $\mathcal{V}_1 = 16 \times 16 \times 32$, $\mathcal{V}_2 = 8 \times 8 \times 16$, $D_1 = 256$, and $D_2 = 512$. 
In the motion encoding path, we sample the combing motion into up to $K = 16$ uniformly spaced key tool poses and convert them into a volumetric grid spatially aligned with the state, as introduced in~\cref{sec:dynamics-phase2}. A contact radius of 3 voxels is used to define the cutoff, forming the local contact region prior. The resulting grid is encoded into control embeddings $c_{\text{bottom}} \in \mathbb{R}^{\mathcal{V}_1 \times D_1}$ and $c_{\text{top}} \in \mathbb{R}^{\mathcal{V}_2 \times D_2}$, which are then fused with $h_{\text{bottom}}$ and $h_{\text{top}}$ using a 3D attention-based feature fusion mechanism for latent state editing. 
The fused embeddings are projected to $d = 32$ and quantized with codebooks of 2048 entries at both levels. The decoders progressively up-sample and decode the quantized embeddings into the original resolution $\mathcal{V}_0$ for end state prediction. To better recover geometric details from the latent, we use transposed convolutions with residual connections instead of direct interpolation for up-sampling.

\textbf{PC-GNN Baseline.}
Our implementation is built on the original RoboCraft framework~\citep{shi2022robocraft}. In this method, computational cost and model capacity are jointly determined by the point cloud resolution, the maximum number of neighbors used to construct edges in the graph, and the number of message passing steps per iteration. To balance these factors while ensuring sufficient message passing capability for capturing hair deformation, which requires dense propagation, we set the maximum number of neighbors to 10, the number of message passing steps to 10, and the point cloud resolution to 2000, which is the highest feasible within the computational budget. The cylindrical tool is represented by 10 points, which are included in the graph together with the hair points.
To enable the model to predict per-point local strand orientation for hair modeling, we extend the original prediction head to output both per-point orientations in the end state and point positions. 
Supervision is applied to both position and orientation predictions. For positions, we use Chamfer Distance between the predicted and ground-truth point clouds. For orientations, we use the bidirectional cosine distance between the predicted and ground-truth orientations, accounting for orientation symmetry, with point-level assignment determined by nearest neighbors. We do not use Earth Mover’s Distance for supervision, as it is computationally expensive and infeasible for training at the point cloud resolution required in our hair scenario.
Following the standard setting of this method, we do not use pre-training for \textbf{PC-GNN}.

\textbf{V-UNet Baseline.}  
We use the same state representation and volumetric motion representation as in the proposed method for \textbf{V-UNet}. The model takes the initial state and motion representation as input and employs a 4-level UNet architecture to progressively encode and decode them for end state prediction. The resolution is down-sampled from $64 \times 64 \times 128$ to $4 \times 4 \times 8$ in the UNet. In each down-sampling and encoding layer (except the first, where the resolution is too high and may cause out-of-memory issues), the motion embedding is fused with the state embedding using a fine-grained, 3D attention-based mechanism, enabling motion information propagation within the grid. During decoding, both state and motion embeddings from the encoder are passed through skip connections. As in the proposed method, we use transposed convolutions with residual connections for up-sampling, and apply the same loss functions for supervision. 
We do not use pre-training for \textbf{V-UNet}.

\textbf{V-FiLM Baseline.}  
\textbf{V-FiLM} builds on the same pre-trained components as the proposed method, including the state encoders, latent codebooks for quantization, and decoders. The key difference is that \textbf{V-FiLM} employs a FiLM~\citep{perez2018film}-based method to adapt the pre-trained model for dynamics modeling.  
Specifically, it encodes the volumetric motion representation into a 1-D latent embedding, then injects it into the two hierarchical state encoders. This conditions the state encoding on the motion signal through a learnable, feature-wise affine transformation applied per channel. After fusion, the embeddings are quantized and decoded to predict the end state, following the same procedure as in the proposed method.  
Unlike the proposed method, which performs fine-grained, spatially aligned state-action fusion, \textbf{V-FiLM} adopts a more straightforward fusion mechanism to incorporate motion signals. We apply the same loss functions as the proposed method for supervision.

\subsection{Data Processing}

To prepare the synthetic hair-combing dynamics data for training, we first convert the 2K strands in each state into dense point clouds with per-point orientations, computed from the ground-truth strand geometry. The point cloud is then down-sampled to 50K points and voxelized into a $64 \times 64 \times 128$ volumetric grid for all voxel-based methods, and to 2K points for \textbf{PC-GNN}. Considering the hair scale in both simulation and the real world, each voxel in the grid corresponds to approximately 5\,mm in the real-world physical space.

\subsection{Evaluation Metric Definition}

For fair comparison, we convert the outputs of all volumetric methods into point clouds with per-point orientations and down-sample them to the same resolution as \textbf{PC-GNN}. Specifically, voxel centers are used as points, with their corresponding voxel orientations assigned as point orientations. The resulting point cloud is then down-sampled to a 2K-point resolution. The ground-truth data is also down-sampled to the same resolution.

We use three metrics for evaluation:

a) \textbf{Point-level Chamfer Distance $\mathrm{CD}_\mathrm{point}$.}  
To evaluate positional accuracy, we compute the symmetric Chamfer Distance between the predicted point cloud $P$ and the ground truth $Q$, ignoring orientation:
{\small
\begin{equation*}
\begin{aligned}
\mathrm{CD}_{\text{point}}(P, Q) = \tfrac{1}{2} \Bigg(
& \frac{1}{|Q|} \sum_{q \in Q} \min_{p \in P} \|q - p\|_2 \\
& + \frac{1}{|P|} \sum_{p \in P} \min_{q \in Q} \|p - q\|_2
\Bigg).
\end{aligned}
\end{equation*}
}%
Note that, in addition to the mean values, we report percentile statistics to better capture the error distributions. This highlights sparse but significant deformations (\eg, affecting only a few strands) that may be obscured by averaging.  
Specifically, for each predicted and ground-truth point, we record the nearest distance to the opposite set, yielding two error distributions. Percentiles are computed for each, and the final percentile values are obtained by averaging over both directions.

b) \textbf{Point-level Orientation Error $\mathrm{Err}_\mathrm{ori}$.}  
To evaluate orientation accuracy, we compute the angular difference between predicted and ground-truth orientations. Each predicted point is matched to its nearest ground-truth neighbor within a positional error threshold of 2\,cm, and their orientation vectors are compared. 
The per-point error is defined as the absolute angular deviation (in degrees) between the two orientations, computed from cosine similarity while accounting for orientation symmetry.
$\mathrm{Err}_{\text{ori}}$ is given by the average error across all predicted points with valid ground-truth assignments within the threshold.
As with $\mathrm{CD}_\mathrm{point}$, we report both mean errors and percentile statistics to capture the distribution of orientation deviations, highlighting sparse but significant deformations.

c) \textbf{Strand-level Chamfer Distance $\mathrm{CD}_\mathrm{strand}$.} 
As a complement to point-level evaluation, we reconstruct strand segments from the point clouds with orientations and assess prediction quality at the strand level. 
This metric evaluates on average how well each predicted strand segment aligns with its closest ground-truth strand segment and vice versa, considering the piecewise-linear curve structure of hair strands.
Formally, for two strand segments $A = \{a_{1}, \dots, a_{M}\}$ and $B = \{b_{1}, \dots, b_{N}\}$,  
the directed distance from $A$ to $B$ is defined as

{\small
\[
D(A, B) = \frac{1}{M} \sum_{k=1}^{M} \min_{\,1 \le l < N} \;\min_{x \in [b_{l}, b_{(l+1)}]} \|a_{k} - x\|_2 ,
\]
}
where $[b_{l}, b_{(l+1)}]$ denotes the line segment between consecutive points of strand $B$.
The strand-level Chamfer Distance is then defined by averaging over both directions: 
{\small
\begin{equation*}
\begin{aligned}
\mathrm{CD}_{\mathrm{strand}}(\mathcal{P}, \mathcal{G}) = \tfrac{1}{2} \Bigg(
& \frac{1}{|\mathcal{P}|} \sum_{P \in \mathcal{P}} \min_{G \in \mathcal{G}} D(P, G) \\
& + \frac{1}{|\mathcal{G}|} \sum_{G \in \mathcal{G}} \min_{P \in \mathcal{P}} D(G, P)
\Bigg),
\end{aligned}
\end{equation*}
}%
where $\mathcal{P}$ and $\mathcal{G}$ denote the sets of predicted and ground-truth strand segments, respectively.
As with $\mathrm{CD}_\mathrm{point}$ and $\mathrm{Err}_\mathrm{ori}$, we report both mean errors and percentile statistics to capture the distribution of strand-level distances, highlighting sparse but significant deformations.

\section{Visual Goal-conditioned Styling Implementation Details}

\subsection{Perception}  
In simulation, we use oracle strand information for all systems to ensure a fair comparison of their visual goal-conditioned planning capabilities.  
Specifically, \textbf{Rule-based System} receives a front-view 2D oracle strand orientation map from the simulator, while all other methods are provided with 3D oracle dense point clouds with per-point orientations.  
In real-world experiments, \textbf{Rule-based System} captures a front-view RGB-D image at 1280$\times$720 resolution using the wrist-mounted camera and estimates a 2D orientation map from it.  
\systemname{} performs 3D state estimation by capturing RGB-D observations at 1280$\times$720 resolution from 56--65 viewpoints around the hair, estimating 2D orientation maps for each, and reconstructing the 3D hair state following~\citet{nam_2019_strandaccurate}.  
For both systems, we use a long exposure time to obtain clearer images for more accurate orientation estimation.
For both systems, we estimate depth maps using an external module based on FoundationStereo~\citep{wen2025foundationstereo}, which computes disparity maps from stereo infrared images and estimates depths, instead of relying on the on-board depth estimation as it typically provides poor performance in our setting. 
For both systems, we obtain hair segmentation masks using Grounded-SAM-2~\citep{ravi2024sam2segmentimages, liu2023grounding, ren2024grounding, ren2024grounded, kirillov2023segany, jiang2024trex2}.

\subsection{Planning}
For all model-based systems, at each step, the planner samples 48 action candidates,~\ie, 3D tool motions, and optimizes an action trajectory using MPPI to minimize the cost. The planning cost is defined as the strand-level distance between strand segments reconstructed from the predicted future states and the goal state. The planner optimizes over a two-step prediction horizon and executes the first step.
For~\textbf{Rule-based System}, we reimplemented the planner ourselves since the original code is not publicly available. As only limited technical details are provided in their paper~\citep{kim_2025_fronthairstyle}, we reproduced their design to the best of our understanding, following their high-level methodology. Specifically, at each step, the planner computes a 2D orientation difference map between the estimated orientations of the current and goal states, samples 300 2D points based on the differences for backward strand extraction and root-centric target strand extraction, derives the 2D tool motion, and converts it back to 3D for execution.

\section{Language Goal-conditioned Styling Implementation Details}

\subsection{Intent Grounding Module}
The intent grounding module makes three model calls: pre-generation reasoning, goal image generation, and post-generation candidate selection. The full prompt for each is given below. Terms in braces are substituted at run time; terms in angle brackets are placeholders within the requested output format and are not substituted.

\paragraph{Pre-generation Reasoning.}
The VLM (Gemini 3.1 Pro~\citep{gemini31pro2026}) receives the user's instruction and an image of the current observation, together with an introduction of the robot's capabilities, and returns a concrete styling plan.
\begin{promptbox}
\begin{lstlisting}
You are an expert hair manipulation AI system. You are facing a user who provides a language description of their desired hairstyle. Your task is to interpret this description and determine how to achieve it.

You can comb the hair freely to reach the desired style. However, you cannot change any intrinsic properties such as hair color, texture, or volume.

You will be given:
1. A language description of the desired hairstyle.
2. An image of the current hair state.

Important:
You are facing the user. Any references to "left" or "right" in the language description are from the user's perspective.

Based on them, you need to:
1. Describe the current hair state in detail;
2. Analyze how the current hair differs from the desired hair style;
3. Plan a concise sequence of combing actions. Each action should be a single, clear instruction using only your perspective, not the user's perspective.

Your answer should be structured as a json object following this format:
{
  "current_hair_state": "<detailed description of the current hair state>",
  "difference_analysis": "<analysis of how the current hair state differs from the desired style>",
  "action_plan": "<planned concise combing actions to achieve the desired hair style>"
}

The language description is: "{user_goal_instruction}"
\end{lstlisting}
\end{promptbox}

\paragraph{Goal Image Generation.}
The image generation model (Gemini 3 Pro Image~\citep{gemini3proimage2025}) is conditioned on the current observation and the styling plan produced above, \ie, the \texttt{action\_plan} field of the reasoning output, rather than on the raw instruction. We sample three candidate goal images in parallel.
\begin{promptbox}
\begin{lstlisting}
{styling_plan} Do not change the hair color, texture, volume, etc.
\end{lstlisting}
\end{promptbox}

\paragraph{Post-generation Candidate Selection.}
The same VLM receives the three candidates alongside the observation and instruction, and returns the index of the candidate most consistent with the user's intent and the robot's capabilities.
\begin{promptbox}
\begin{lstlisting}
You are an expert hair manipulation AI system. You are facing a user who provides a language description of their desired hairstyle. Your task is to interpret this description and determine the most appropriate achievable result.

You can comb the hair freely to reach the desired style. However, you cannot change any intrinsic properties such as hair color, texture, or volume.

You will be given:
1. A language description of the desired hairstyle.
2. An image of the current hair state.
3. Three candidate images representing possible final hairstyles (candidate 0, candidate 1, candidate 2).

Important:
You are facing the user. Any references to "left" or "right" in the language description are from the user's perspective.

Based on them, you need to evaluate all three candidate images and select the one that best satisfies the following criteria:
1. Accurately reflects the desired hairstyle described in the language goal.
2. Is realistically achievable from the current hair state using only combing actions.
3. Preserves intrinsic properties such as hair color, texture, and volume.

You must:
1. Analyze each candidate separately.
2. Keep each analysis brief, precise, and comparable across candidates.
3. Avoid redundancy and avoid mixing multiple interpretations.

Your answer should be structured as a json object following this format:
{
    "analysis": {
        "candidate_0": "<brief analysis of how well candidate 0 satisfies the criteria>",
        "candidate_1": "<brief analysis of how well candidate 1 satisfies the criteria>",
        "candidate_2": "<brief analysis of how well candidate 2 satisfies the criteria>"
    },
    "selected_index": <index of the selected best candidate, either 0, 1, or 2>
}

The language description is: "{user_goal_instruction}"
\end{lstlisting}
\end{promptbox}

\subsection{Model-based Planning}
Planning in the language branch follows the visual branch except in the cost function: the same dynamics model, MPPI planner, action space, planning horizon, and replanning schedule are used throughout. The difference is that the goal is a 2D image rather than a 3D volumetric state, so the cost is evaluated in the image plane.

At each MPPI iteration, the 3D hair state predicted by the dynamics rollout is projected onto the image plane using the calibrated camera parameters, producing a predicted hair mask $\hat{M}$. The goal image is likewise converted to a binary hair mask $M^{*}$. The cost is the symmetric Chamfer Distance between the two masks' pixel sets:
{\small
\begin{equation*}
\begin{aligned}
\mathrm{CD}_{\text{pixel}}(\hat{M}, M^{*}) = \tfrac{1}{2} \Bigg(
& \frac{1}{|\hat{M}|} \sum_{q \in \hat{M}} \min_{p \in M^{*}} \|q - p\|_2 \\
& + \frac{1}{|M^{*}|} \sum_{p \in M^{*}} \min_{q \in \hat{M}} \|p - q\|_2
\Bigg),
\end{aligned}
\end{equation*}
}%
where $q$ and $p$ index the occupied pixels of the predicted and goal masks respectively.

\subsection{Baselines in the User Study}

The Study 1 ablations differ from the full grounding module in which mediation stages they retain, and the Study 2 baselines differ from~\systemname{} in how actions are planned and selected.

\paragraph{Study 1: intent grounding ablations.}

\textbf{Direct Grounding} removes both mediation stages: the image generation model is directly conditioned on the current observation and the raw instruction, with no pre-generation reasoning and no candidate selection, and the single generated image is used as the goal. Its generation prompt is therefore the goal image generation prompt above with \texttt{\{styling\_plan\}} replaced by \texttt{\{user\_goal\_instruction\}}, \ie, the raw instruction from the user.

\textbf{Reasoned Grounding} removes only candidate selection: pre-generation reasoning is run as in the full module and the resulting styling plan conditions the generation model, but a single image is sampled and used directly as the goal.

\paragraph{Study 2: planning baselines.}
\textbf{VLM Planner} replaces the MPPI planner while keeping the full grounding module. At each step the VLM receives the current observation, the language instruction, and the grounded goal image, together with a set of sampled candidate actions annotated on the observation following PIVOT~\citep{nasiriany2024pivot}, and returns the index of the action it judges most likely to advance the hair toward the goal.
Candidate actions are sampled in two stages. In the first stage, we overlay a grid on the hair region of the observation and ask the VLM to select which cells can serve as start and end regions for a combing motion. Specifically, we ask the VLM to select 2 start cells and, for each, 2 end cells, giving 4 candidate start-end region pairs (\texttt{top\_n\_start} and \texttt{top\_n\_end} are both set to 2).

\begin{promptbox}
\begin{lstlisting}
You are an expert hair manipulation AI system. You are facing a user who provides a goal image of their desired hairstyle. Your task is to determine how to achieve this hairstyle through combing actions.

You can comb the hair freely to reach the desired style.

You will be given:
1. An image of the current hair state.
2. An image of the desired hairstyle (goal image).
3. An image of the current hair state overlaid with a red grid. Each grid cell has a unique red ID.

Based on these inputs, you need to:
1. Compare the current hair state with the goal hairstyle and identify the key differences.
2. Select the top {top_n_start} grid cells as starting points for combing motions.
3. For each selected start grid, select the top {top_n_end} grid cells as end points of the combing motion.

Selection Requirements:
1. Chosen grid pairs should directly address the differences between the current and goal hairstyles.

Your answer should be structured as a json object following this format:
{
    "analysis": "<concise explanation of key differences and how they guide the selected motions>",
    "chosen_grid_pairs": [
        {"start_grid_id": <start_grid_id>, "end_grid_id": [<end_grid_id_1>, <end_grid_id_2>, ...]},
        ...
    ]
}
\end{lstlisting}
\end{promptbox}

In the second stage, we sample 10 concrete actions within the selected regions to form the candidate set for PIVOT-style visual prompting. We annotate them on the observation as arrows with unique IDs.

\begin{promptbox}
\begin{lstlisting}
You are an expert hair manipulation AI system. You are facing a user who provides a goal image of their desired hairstyle. Your task is to determine how to achieve this hairstyle through combing actions.

You can comb the hair freely to reach the desired style.

You will be given:
1. An image of the current hair state.
2. An image of the desired hairstyle (goal image).
3. An image of the current hair state overlaid with green arrows representing candidate combing motions. Each motion has a unique red ID.

Based on these inputs, you need to:
1. Evaluate each candidate motion by how effectively it moves the current hair state toward the goal hairstyle.
2. Determine which single motion best addresses the most important differences between the current and goal hairstyles.

Selection Requirements:
1. Focus on impact: prefer motions that produce meaningful visible changes toward the goal.
2. Do not select motions that cause minimal change or are irrelevant to the goal.

Your answer should be structured as a json object following this format:
{
    "evaluation": "<concise explanation of how each motion addresses the differences and why the best motion is chosen>",
    "best_motion_id": <best_motion_id>
}
\end{lstlisting}
\end{promptbox}

\textbf{End-to-End VLM} uses the same two-stage selection procedure but receives no grounded goal image, acting directly from the observation and the language instruction. The prompts follow those of \textbf{VLM Planner} with the goal image removed from the input list and replaced by the language instruction, and with references to comparing against the goal image restated as comparing against the hairstyle the instruction asks for. As in the intent grounding prompts, the first stage additionally states that left and right in the description are from the user's perspective, since the instruction is now interpreted directly rather than through a grounded image.

\textbf{PC-GNN Model} keeps \systemname{}'s intent grounding module and MPPI planner unchanged, and substitutes only the dynamics model the planner queries during rollouts, replacing the volumetric model with the PC-GNN baseline of~\cref{sec:exp-dynamics-baseline}.

\end{document}